\AddToHook{package/lineno/before}{\RequirePackage{amsmath}}
\documentclass{ecai}
\usepackage{times}

\usepackage{algorithm}
\usepackage[noend]{algorithmic}

\usepackage{tabularx}
\usepackage{xparse}
\usepackage{amsmath}
\usepackage{amsthm}
\usepackage{amssymb}
\usepackage{amsfonts}
\usepackage{xspace}
\usepackage{relsize}
\usepackage{graphicx}
\usepackage{booktabs}
\usepackage{multirow}
\usepackage{diagbox}
\usepackage{cancel}
\usepackage{comment}
\usepackage[export]{adjustbox}
\usepackage[svgnames]{xcolor}
\usepackage{paralist}

\usepackage{xr}

\usepackage{bm}
\usepackage{alphabeta}          
\usepackage{stmaryrd}

\def\eqref#1{equation~\ref{#1}}

\def\1{\bm{1}}
\def\0{\bm{0}}

\def\rh{{\textnormal{h}}}

\def\rr{{\textnormal{r}}}

\def\rt{{\textnormal{t}}}

\def\rx{{\textnormal{x}}}

\def\rS{{\textnormal{S}}}

\def\rX{{\textnormal{X}}}

\DeclareMathAlphabet{\mathsfit}{\encodingdefault}{\sfdefault}{m}{sl}
\SetMathAlphabet{\mathsfit}{bold}{\encodingdefault}{\sfdefault}{bx}{n}

\def\E{{\mathbb{E}}}

\def\N{{\mathcal{N}}}

\def\R{{\mathbb{R}}}

\def\Z{{\mathbb{Z}}}

\DeclareMathOperator*{\argmin}{arg\,min}

\newcommand{\brackets}[1]{{\left[#1\right]}}
\newcommand{\braces}[1]{{\left\{#1\right\}}}
\newcommand{\parens}[1]{{\left(#1\right)}}

\newcommand{\dbrackets}[1]{{\left\llbracket#1\right\rrbracket}}

\NewDocumentCommand{\diffby}{s m O{}}{
 \IfBooleanTF{#1}
  {\frac{\partial#3}{\partial#2}}
  {\frac{d#3}{d#2}}
}

\newcommand{\satisfies}{\vDash}

\RenewDocumentCommand{\to}{o o}{
 \IfNoValueTF{#1}
  {\rightarrow}
  {\IfNoValueTF{#2}
   {\xrightarrow{#1}}
   {\xrightarrow[#2]{#1}}}
}

\NewDocumentCommand{\affect}{o o}{
 \IfNoValueTF{#1}
  {\rightsquigarrow}
  {
   \IfNoValueTF{#2}
   {\rightsquigarrow^{#1}}
   {\rightsquigarrow^{#1}_{#2}}
  }
}

\renewcommand{\iff}{\Leftrightarrow}

\newcommand{\erf}{\function{erf}}

\newcommand{\iid}{i.i.d.\xspace}

\usepackage{stackengine}
\stackMath
\newcommand\tsup[2][2]{
 \def\useanchorwidth{T}
  \ifnum#1>1
    \stackon[-.5pt]{\tsup[\numexpr#1-1\relax]{#2}}{\scriptscriptstyle\sim}
  \else
    \stackon[.5pt]{#2}{\scriptscriptstyle\sim}
  \fi
}

\newcommand{\function}[1]{\textsc{#1}}

\def\_{\\[-0.3em]}

\newtheorem{defi}{Definition}

\newtheorem{theo}{Theorem}

\newtheorem{corollary}{Corollary}

\makeatletter
\let\@myref\ref

\newcommand{\refsec}[1]{Sec.\,\@myref{#1}}
\newcommand{\refseq}[1]{Sec.\,\@myref{#1}}
\newcommand{\refig}[1]{Fig.\,\@myref{#1}}
\newcommand{\reftbl}[1]{Table\,\@myref{#1}}
\newcommand{\refstep}[1]{Step \@myref{#1}}
\newcommand{\refalgo}[1]{Alg.\,\@myref{#1}}
\newcommand{\refchap}[1]{Chap.\,\@myref{#1}}
\newcommand{\reflst}[1]{List \@myref{#1}}
\newcommand{\refeq}[1]{Eq.\,\@myref{#1}}

\newcommand{\reftheo}[1]{Thm.\,\@myref{#1}}
\newcommand{\refline}[1]{line\,\@myref{#1}}
\newcommand{\refdef}[1]{Def.\, \@myref{#1}}
\newcommand{\refex}[1]{Example\,\@myref{#1}}
\newcommand{\refconv}[1]{Conv.\,\@myref{#1}}
\newcommand{\reffact}[1]{Fact.\,\@myref{#1}}

\newcommand{\refsecs}[2]{Sec.\,\@myref{#1}-\@myref{#2}}
\newcommand{\refseqs}[2]{Sec.\,\@myref{#1}-\@myref{#2}}
\newcommand{\refigs}[2]{Fig.\,\@myref{#1}-\@myref{#2}}
\newcommand{\reftbls}[2]{Tables \@myref{#1}-\@myref{#2}}
\newcommand{\refsteps}[2]{Steps \@myref{#1}-\@myref{#2}}
\newcommand{\refalgos}[2]{Alg.\,\@myref{#1}-\@myref{#2}}
\newcommand{\refchaps}[2]{Chap.\,\@myref{#1}-\@myref{#2}}
\newcommand{\reflsts}[2]{Lists \@myref{#1}-\@myref{#2}}
\newcommand{\refeqs}[2]{Eq.\,\@myref{#1}-\@myref{#2}}
\newcommand{\refpages}[2]{p.\pageref{#1}-\@myref{#2}}
\newcommand{\reftheos}[2]{Thm.\,\@myref{#1}-\@myref{#2}}
\newcommand{\reflines}[2]{line\,\@myref{#1}-\@myref{#2}}
\newcommand{\refdefs}[2]{Def.\,\@myref{#1}-\@myref{#2}}
\newcommand{\refexs}[2]{Example\,\@myref{#1}-\@myref{#2}}
\newcommand{\refconvs}[2]{Conv.\,\@myref{#1}-\@myref{#2}}
\newcommand{\reffacts}[2]{Facts.\,\@myref{#1}-\@myref{#2}}

\makeatother

\newcounter{list}[section]

\makeatletter
\@ifundefined{citet}{
  \NewDocumentCommand{\citet}{o m}{
    \IfNoValueTF{#1}
      {\citeauthor{#2} (\citeyear{#2})}
      {\citeauthor{#2} (\citeyear[#1]{#2})}
  }
}{}
\@ifundefined{citep}{
  \NewDocumentCommand{\citep}{o m}{
    \IfNoValueTF{#1}
      {\cite{#2}}
      {\cite[#1]{#2}}
  }
}{}
\makeatother

\newlength{\maxwidth}

\newcommand{\pre}{\function{pre}}
\newcommand{\adde}{\function{add}}
\newcommand{\dele}{\function{del}}

\newcommand{\cost}{\function{c}}

\def\hash{\text{\relsize{-1}\#}}
\newcommand{\ar}[1]{\hash{}#1}

\newcommand{\lsota}{state-of-the-art\xspace}  

\newcommand{\astar}{\ifmmode{A^*}\else{A$^*$}\fi\xspace}
\newcommand{\gbfs}{\ifmmode{\mathrm{GBFS}}\else{GBFS}\fi\xspace}
\NewDocumentCommand{\uct}{s}{\ifmmode{\mathrm{UCT}{\IfBooleanT{#1}{^*}}}\else{UCT{\IfBooleanT{#1}{*}}}\fi\xspace}
\NewDocumentCommand{\guct}{s}{\ifmmode{\mathrm{GUCT}{\IfBooleanT{#1}{^*}}}\else{GUCT{\IfBooleanT{#1}{*}}}\fi\xspace}

\newcommand{\topen}{tree-based open list\xspace}

\makeatletter

\newcommand{\newheuristic}[2]{
 \def#1{
  \relax\ifmmode
  h^\mathrm{#2}\xspace
  \else
  \text{#2}\xspace
  \fi
 }
}

\newheuristic{\lmcut}{LMcut}
\newheuristic{\mands}{M\&S}
\newheuristic{\pdb}{PDB}
\newheuristic{\ff}{FF}
\newheuristic{\ce}{CEA}
\newheuristic{\cg}{CG}
\newheuristic{\gc}{GC}
\newheuristic{\ad}{add}
\newheuristic{\hmax}{max}
\newheuristic{\lc}{LC}
\newheuristic{\blind}{blind}

\newcommand{\newlearnedheuristic}[2]{
 \def#1{
  \relax\ifmmode
  H^\mathrm{#2}\xspace
  \else
  \text{#2}\xspace
  \fi
 }
}

\newlearnedheuristic{\Hlmcut}{LMcut}
\newlearnedheuristic{\Hmands}{M\&S}
\newlearnedheuristic{\Hpdb}{PDB}
\newlearnedheuristic{\Hff}{FF}
\newlearnedheuristic{\Hce}{CEA}
\newlearnedheuristic{\Hcg}{CG}
\newlearnedheuristic{\Had}{add}
\newlearnedheuristic{\Hmax}{max}
\newlearnedheuristic{\Hlc}{LC}
\newlearnedheuristic{\Hblind}{blind}

\newcommand{\newUnitCostHeuristic}[2]{
 \def#1{
  \relax\ifmmode
  \hat{h}^\mathrm{#2}\xspace
  \else
  \text{#2}\xspace
  \fi
 }
}

\newUnitCostHeuristic{\lmcuto}{LMcut}
\newUnitCostHeuristic{\mandso}{M\&S}
\newUnitCostHeuristic{\ffo}{FF}
\newUnitCostHeuristic{\ceo}{CEA}
\newUnitCostHeuristic{\cgo}{CG}
\newUnitCostHeuristic{\ado}{add}
\newUnitCostHeuristic{\gco}{GoalCount}
\newUnitCostHeuristic{\lco}{LC}

\newcommand{\newrandomheuristic}[2]{
 \def#1{
  \ifmmode
  \rh^\mathrm{#2}\xspace
  \else
  \text{#2}\xspace
  \fi
 }
}

\newrandomheuristic{\rlmcut}{LMcut}
\newrandomheuristic{\rmands}{M\&S}
\newrandomheuristic{\rpdb}{PDB}
\newrandomheuristic{\rff}{FF}
\newrandomheuristic{\rce}{CEA}
\newrandomheuristic{\rcg}{CG}
\newrandomheuristic{\rad}{add}
\newrandomheuristic{\rhmax}{max}
\newrandomheuristic{\rlc}{LC}

\makeatother

\usepackage{xkeyval}
\makeatletter

\define@key{strips}{negative-preconditions}[1]{\def\@conditiontype{#1}}
\define@key{strips}{disjunctive-preconditions}[2]{\def\@conditiontype{#1}}
\define@key{strips}{conditional-effects}[1]{\def\@usecondeffect{#1}}
\define@key{strips}{action-costs}[1]{\def\@usecost{#1}}
\define@key{strips}{derived-predicates}[1]{\def\@useaxiom{#1}}
\define@key{strips}{lifted}[1]{\def\@lifted{#1}}
\define@key{strips}{optimal}[1]{\def\@optimal{#1}}
\define@key{strips}{unitcost}[1]{\def\@unitcost{#1}}

\def\conditionset{
\if\@useaxiom0
P
\else
P\cup P_X
\fi
}

\let\satisfies@orig\satisfies

\def\satisfies{
\if\@conditiontype0
\supseteq
\else
\satisfies@orig
\fi
}

\def\condition{
\if\@conditiontype0
\conditionset
\else
\mathcal{F}(\conditionset)
\fi
}

\def\ga{
\if\@lifted0
a
\else
a^{\dagger}
\fi
}

\def\applyformula{
\if\@usecondeffect0
(s \setminus \dele(a)) \cup \adde(a)
\else
(s
 \setminus \braces{e \mid (c \triangleright e) \in \dele(\ga), c\satisfies s})
 \cup      \braces{e \mid (c \triangleright e) \in \adde(\ga), c\satisfies s}
\fi
}

\NewDocumentCommand{\strips}{O{}}{
\def\@useaxiom{0}
\def\@conditiontype{0}
\def\@usecondeffect{0}
\def\@usecost{0}
\def\@lifted{0}
\def\@optimal{0}
\def\@unitcost{0}
\setkeys{strips}{#1}
\if\@lifted1
  \strips@propositional\par
  \strips@lifted
\else
  \strips@propositional
\fi
}

\newcommand{\strips@propositional}{
\if\@conditiontype1
Given a set of propositional variables $V$,
let $\mathcal{F}(V)$ be a propositional formula consisting of $V$ and
logical operations $\braces{\land,\lnot}$.
\fi
\if\@conditiontype2
Given a set of propositional variables $V$,
let $\mathcal{F}(V)$ be a propositional formula consisting of $V$ and
logical operations $\braces{\land,\lor,\lnot}$.
\fi
\if\@useaxiom0
We define a propositional STRIPS Planning problem
as a 4-tuple $\brackets{P,A,I,G}$
where
 $P$ is a set of propositional variables,
 $A$ is a set of actions,
 $I\subseteq P$ is the initial state, and
 $G\subseteq \conditionset$ is a goal condition.
\else
We define a propositional STRIPS Planning problem
as a 6-tuple $\brackets{P,A,X,P_X,I,G}$
where
 $P$ is a set of propositions,
 $A$ is a set of actions,
 $X$ is a set of axioms,
 $P_X$ is a set of derived propositions ($P\cap P_X=\emptyset$),
 $I\subset P$ is the initial state, and
 $G\subset \conditionset$ is a goal condition.
\fi
\if\@usecost0
Each action $a\in A$ is a 3-tuple $\brackets{\pre(a),\adde(a),\dele(a)}$ where
\else
Each action $a\in A$ is a 4-tuple $\brackets{\pre(a),\adde(a),\dele(a),\cost(a)}$ where
$\cost(a) \in \Z^{0+}$ is a cost,
\fi
$\pre(a) \subseteq \condition$ is a precondition and
\if\@usecondeffect0
$\adde(a), \dele(a)\subseteq P$ are the add-effects and delete-effects.
\else
$\adde(a), \dele(a)$ are the add-effects and delete-effects.
Each effect is denoted as $c \triangleright e$ where
$c \in \condition$ is an \emph{effect condition} and
$e \in P$.
\fi
\if\@useaxiom1
The set of axioms $X$ consists of clauses $f \Rightarrow p$ where
$f \in \condition$ is a body and $p \in P_X$ is a head.
\fi
A state $s\subseteq \conditionset$ is a set of true propositions
(all of $P\setminus s$ is false),
an action $a$ is \emph{applicable} when $s \satisfies \pre(a)$ (read: $s$ \emph{satisfies} $\pre(a)$),
and applying action $a$ to $s$ yields a new successor state
\if\@useaxiom0
$a(s) = \applyformula$.
\else
$a(s)$.
To compute $a(s)$, we first obtain a non-derived state
$a'(s) = \applyformula \setminus P_X $.
Then we perform a fix-point calculation such that
$s \gets s \cup \braces{p \in P_X \mid (f\Rightarrow p)\in X \land s \satisfies f}$
where $s$ is initialized to $a'(s)$.
\fi

The task of classical planning is to find a sequence of actions called a \emph{plan} $(\ga_1,\cdots,\ga_n)$
where, for $1\leq t\leq n$,
 $s_0=I$, $s_t\satisfies \pre(a_{t+1})$, $s_{t+1}=a_{t+1}(s_t)$,
 and $s_n\satisfies G$.
\if\@optimal1
 A plan is \emph{optimal} if
 \if\@usecost0
   there is no shorter plan.
 \else
   there is no plan with lower cost $\sum_t \cost(a_t)$.
 \fi
 A plan is otherwise called \emph{satisficing}.
 \if\@usecost1
  \if\@unitcost1
  In this paper, we assume unit-cost: $\forall a\in A; \cost(a)=1$.
  \fi
 \fi
\fi
}

\newcommand{\strips@lifted}{
In \emph{Lifted STRIPS}, each propositional variable is an \emph{instantiation}/\emph{grounding} of
a first-order logic predicate.
Each predicate $p(x_1,\ldots,x_{\ar{p}})$ is parameterized by a list of parameters/variables/arguments $X=(x_1,\ldots,x_{\ar{p}})$,
where $\ar{p}$ is an \emph{arity} of $p$.
A proposition is obtained by substituting each $x_i$ with an \emph{object} in a set $O$.
Each $p$ therefore has $O^{\ar{p}}$ instantiations.
Similarly, each action $a\in A$ is now called a \emph{ground action},
which is an instantiation of a \emph{lifted action} $a(x_1,\ldots,x_{\ar{p}})$ parameterized by $\ar{a}$ parameters.
A ground action is obtained by substituting the arguments as well as
the parameters used in the preconditions and the effects.
}

\makeatother

\allowdisplaybreaks

\begin{document}

\begin{frontmatter}

\paperid{1554}

\title{Scale-Adaptive Balancing of Exploration and Exploitation in Classical Planning}

\author[A]{\fnms{Stephen}~\snm{Wissow}\thanks{Corresponding Author. Email: swj@unh.edu}\footnote{Equal contribution.}}
\author[B]{\fnms{Masataro}~\snm{Asai}\footnotemark}

\address[A]{University of New Hampshire}
\address[B]{MIT-IBM Watson AI Lab}

\begin{abstract}
Balancing exploration and exploitation has been an important problem in
both adversarial games and automated planning.
While it has been extensively analyzed in the Multi-Armed Bandit (MAB) literature,
and the game community has achieved great success with MAB-based Monte Carlo Tree Search (MCTS) methods,
the planning community has struggled to advance in this area.
We describe how Upper Confidence Bound 1's (UCB1's) assumption of reward distributions with known bounded support shared among siblings (arms) is violated when MCTS/Trial-based Heuristic Tree Search (THTS) in previous work uses heuristic values of search nodes in classical planning problems as rewards.
To address this issue,
we propose a new Gaussian bandit, UCB1-Normal2, and analyze its regret bound.
It is variance-aware like UCB1-Normal and UCB-V, but has a distinct advantage:
it neither shares UCB-V's assumption of known bounded support nor relies on UCB1-Normal's conjectures on Student's $t$ and $\chi^2$ distributions.
Our theoretical analysis predicts that UCB1-Normal2 will perform well when the estimated variance is accurate,
which can be expected in deterministic, discrete, finite state-space search, as in classical planning.
Our empirical evaluation confirms that
MCTS combined with UCB1-Normal2 outperforms Greedy Best First Search (traditional baseline)
as well as MCTS with other bandits.
\end{abstract}

\end{frontmatter}

\section{Introduction}

From the early history of AI and in particular of automated planning and scheduling,
heuristic forward search has been a primary methodology for tacking challenging combinatorial problems.
A rich variety of search algorithms have been proposed, including Dijkstra search \citep{dijkstra1959note},
\astar / WA$^*$ \citep{hart1968formal}, and Greedy Best First Search \citep[GBFS]{bonet2001planning}.
They are divided into three categories:
\emph{optimizing}, which must guarantee the optimality of the output,
\emph{satisficing}, which may or may not attempt to minimize solution cost,
and \emph{agile}, which ignores solution cost and focuses on finding a solution quickly.
This paper focuses on the agile setting.

Unlike optimizing search,
theoretical understanding of satisficing and agile search has been limited.
Recent theoretical work on GBFS \citep{heusner2017understanding,heusner2018search,heusner2018best,kuroiwa2022biased}
refined the concept of search progress in agile search,
but only based on a post hoc analysis that depends on oracular information,
making their insights difficult to apply to practical search algorithm design,
although it has been recently applied to a learning-based approach \citep{ferber2022learning}.
More importantly,
their analysis is incompatible with a wider range of randomized algorithms
\citep{nakhost2009monte,imai2011novel,kishimoto2012diverse,valenzano2014comparison,xie2012planning,xie14gbfsle,xie14type,xie15understanding,Asai2017b,kuroiwa2022biased}
that outperform the deterministic baseline with randomized explorations;
as a result,
their detailed theoretical properties are largely unknown
except for probabilistic completeness \citep{valenzano2014comparison}.
It is unsurprising that analyzing randomized algorithms requires a statistical perspective,
which is also growing more important
due to recent advances in learned heuristic functions
\citep{toyer2018action,ferber2020study,shen2020learning,ferber2022neural,rivlin2020generalized,gehring2021pddlrl,garrett2016learning}.

In this paper, we tackle the problem of balancing exploration and exploitation in classical planning
through a statistical lens and from the perspective of MABs.
Previous work showed that traditional forward search algorithms (A*, GBFS) can be seen as
a form of MCTS, but we refine and recast this paradigm as
a repeated process of collecting a reward dataset
and exploring the environment based on estimates obtained from this dataset.
This perspective reveals several theoretical issues in GreedyUCT \citep{schulte2014balancing},
a MCTS modified for classical planning that uses UCB1.
Among other things,
the optimization objective of classical planning has no a priori known bound,
which violates the bounded reward assumption of UCB1.

To apply MAB to classical planning correctly,
we propose UCB1-Normal2, a new Gaussian bandit, and
GreedyUCT-Normal2, a new agile planning algorithm that combines MCTS with UCB1-Normal2,
and show that GreedyUCT-Normal2 outperforms
traditional agile algorithms (GBFS),
\lsota diversified search (Softmin-Type(h) \citep{kuroiwa2022biased}),
existing MCTS-based algorithms (GreedyUCT, GreedyUCT*),
MCTS combined with other variance-aware bandits (UCB1-Normal and UCB-V \citep{audibert2009exploration})
or simple-regret bandits (TTTS \citep{russo2020simple}).

While most of our empirical analyses are based on Pyperplan-based implementation
and focus on algorithmic efficiency rather than on low-level performance,
we also re-implemented the algorithms in C++/Fast-Downward
and performed evaluations in IPC 2018 satisficing instances,
where \guct-Normal2 outperformed Softmin-Type(h) on the number of instances solved.

In summary, our core contributions are as follows.
\begin{compactitem}
 \item We identify theoretical issues that arise when applying UCB1 to planning tasks.
 \item To address these issues, we present UCB1-Normal2, a new Gaussian bandit.
       We analyze its regret bound, which improves as the estimated variance is closer to the true variance,
       and is constant when they match.
       This makes it particularly powerful in a deterministic and finite state space such as classical planning.
 \item GreedyUCT-Normal2, a new forward search algorithm that combines UCB1-Normal2 with MCTS,
       outperforms existing algorithms in agile classical planning.
\end{compactitem}

The code for this paper is available at \url{https://github.com/IBM/pyperplan-mcts-public}.

\section{Background}

\label{sec:classical-planning}

\strips[optimal,action-costs,unitcost]

A domain-independent heuristic function $h$ in classical planning is
a function of a state $s$ and the problem $\brackets{P,A,I,G}$,
but the notation $h(s)$ usually omits the latter.
It returns an estimate of the cumulative cost from $s$ to one of the goal states (which satisfy $G$),
typically through a symbolic, non-statistical means including problem relaxation and abstraction.
Notable \lsota functions that appear in this paper include
$\ff, \hmax, \ad$, and $\gc$ \citep{hoffmann01,bonet2001planning,FikesHN72}.
Their implementation details are beyond the scope of this paper, and are included in the appendix \cite[\refsec{sec:heuristics}]{wissow2023scale:full}.

\subsection{Multi-Armed Bandit (MAB)}

MAB \citep{thompson1933likelihood,robbins1952some,bush1953stochastic} is a problem of
finding the best strategy to choose from multiple unknown reward distributions.
It is typically depicted by a row of $K$ slot machines each with a lever or ``arm.''
Each time the player plays one of the machines and pulls an arm (a \emph{trial}),
the player receives a reward sampled from the distribution assigned to that arm.
Through multiple trials, the player discovers the arms' distributions and selects arms to maximize the reward.

The most common optimization objective of MAB is \emph{Cumulative Regret} (CR) minimization.
Let $\rr_i$ ($1\leq i \leq K$) be a random variable (RV) for the reward that we would receive when we pull arm $i$.
We call $p(\rr_i)$ an unknown \emph{reward distribution} of $i$.
Let $\rt_{i}$ be a RV of the number of trials performed on arm $i$ and
$T=\sum_i \rt_{i}$ be the total number of trials across all arms.
\begin{defi}
 The \emph{cumulative regret} $\Delta$ is the gap between the optimal and the actual expected cumulative reward:
 $
 \textstyle\Delta=T\max_i \E[\rr_i] - \sum_i \E[\rt_{i}] \E[\rr_i].
 $
 \label{def:cr}
\end{defi}
Algorithms whose regret per trial $\Delta/T$ converges to 0 with $T\to\infty$
are called \emph{zero-regret}.
Those with a logarithmically upper-bounded regret, $O(\log T)$,
are also called \emph{asymptotically optimal} because
this is the theoretical optimum achievable by any algorithm \citep{lai1985asymptotically}.
Regret bounds tell the speed of convergence, thus its proof is stronger than that of the convergence proof.

Upper Confidence Bound 1 \citep[UCB1]{auer2002finite} is
a logarithmic CR MAB for rewards $\rr_i\in [0,c]$ with a known $c$.
Let $r_{i1}\ldots r_{it_i}\sim p(\rr_i)$ be $t_i$ \iid samples obtained from an arm $i$.
Let $\hat{\mu}_i=\frac{1}{t_i}\sum_{j=1}^{t_i} r_{ij}$.
To minimize CR,
UCB1 selects $i$ with the largest Upper Confidence Bound defined below.
\begin{align}
 \begin{split}
  \text{UCB1}_i &\textstyle= {\hat{\mu}_i + c\sqrt{{2\log T}/{t_i}}}\\
  \text{LCB1}_i &\textstyle= {\hat{\mu}_i - c\sqrt{{2\log T}/{t_i}}}
 \end{split}
 \label{eq:ucb1}
\end{align}
For reward (cost) minimization, LCB1 instead selects $i$ with the smallest Lower Confidence Bound defined above
(e.g., in \citet{kishimoto2022bandit}),
but we may use the terms U/LCB1 interchangeably.
UCB1's second term is often called an \emph{exploration term}.
Generally,
an LCB is obtained by flipping the sign of the exploration term in a UCB.
U/LCB1 refers to a specific algorithm while U/LCB refers to general confidence bounds.
$c$ is sometimes set heuristically as a hyperparameter called the \emph{exploration rate}.

\subsection{Forward Heuristic Best-First Search}
\label{sec:mcts}

Classical planning problems are typically solved as a path finding problem
defined over a state space graph induced by the transition rules,
and the current dominant approach is based on \emph{forward search}.
Forward search maintains a set of search nodes called an \emph{open list}.
They repeatedly
(1) (\emph{selection}) select a node from the open list,
(2) (\emph{expansion}) generate its successor nodes,
(3) (\emph{evaluation}) evaluate the successor nodes, and
(4) (\emph{queueing}) reinsert them into the open list.
Termination typically occurs when a node is expanded that satisfies a goal condition,
but a satisficing/agile algorithm can perform \emph{early goal detection},
which immediately checks whether any successor node generated in step (2) satisfies the goal condition.
Since this paper focuses on agile search, we use early goal detection for all algorithms.

Within forward search,
forward \emph{best-first} search defines a particular ordering in the open list
by defining \emph{node evaluation criteria} (NEC) $f$ for selecting the best node in each iteration.
Let us denote a node by $n$ and the state represented by $n$ as $s_n$.
As NEC,
Dijkstra search uses $f_{\mathrm{Dijkstra}}(n)=g(n)$ ($g$-value), the minimum cost from the initial state $I$ to the state $s_n$ found so far.
\astar uses $f_{\astar}(n)=g(n)+h(s_n)$, the sum of $g$-value and the value returned by a heuristic function $h$ ($h$-value).
GBFS uses $f_{\gbfs}(n)=h(s_n)$.
Forward best-first search that uses $h$ is called forward \emph{heuristic} best-first search.
Dijkstra search is a special case of \astar with $h(s)=0$.

Typically, an open list is implemented as a priority queue ordered by NEC.
Since the NEC can be stateful, e.g., $g(s_n)$ can update its value,
a priority queue-based open list assumes monotonic updates to the NEC
because it has an unfavorable time complexity for removals.
\astar, Dijkstra, and GBFS satisfy this condition because
$g(n)$ decreases monotonically and $h(s_n)$ is constant.

MCTS is a class of forward heuristic best-first search
that represents the open list as the leaves of a tree.
We call the tree a \emph{\topen}.
Our MCTS is based on the description in \citep{keller2013trial,schulte2014balancing}.
Overall, MCTS works in the same manner as other best-first search with a few key differences.
(1) (\emph{selection}) To select a node from the \topen,
it recursively selects an action on each branch of the tree, start from the root, using the NEC
to select a successor node,
descending until reaching a leaf node.
(Sometimes the action selection rule is also called a \emph{tree policy}.)
At the leaf, it
(2) (\emph{expansion}) generates successor nodes,
(3) (\emph{evaluation}) evaluates the new successor nodes,
(4) (\emph{queueing}) attaches them to the leaf, and
\emph{backpropagates} (or \emph{backs-up}) the information to the leaf's ancestors, all the way up to the root.

The evaluation obtains a heuristic value $h(s_n)$ of a leaf node $n$.
In adversarial games like Backgammon or Go, it is obtained either by
(1) hand-crafted heuristics,
(2) \emph{playouts} (or \emph{rollouts})
where the behaviors of both players are simulated
by uniformly random actions (\emph{default policy}) until the game terminates,
or (3) a hybrid \emph{truncated simulation},
which returns a hand-crafted heuristic after performing a short simulation \citep{gelly2011monte}.
In recent work, the default policy is replaced by a learned policy \citep{alphago}.

Trial-based Heuristic Tree Search \citep[THTS]{keller2013trial,schulte2014balancing},
a MCTS for classical planning,
is based on two key observations:
(1) the rollout is unlikely to terminate in classical planning due to sparse goals,
unlike adversarial games, like Go, which are guaranteed to finish in a limited number of steps with a clear outcome (win/loss); and
(2) a \topen can reorder nodes efficiently under non-monotonic updates to NEC, and
thus is more flexible than a priority queue-based open list, and
can readily implement standard search algorithms such as \astar and GBFS without significant performance penalty.
In this paper, we use THTS and MCTS interchangeably.

Finally, Upper Confidence Bound applied to trees \citep[UCT]{kocsis2006bandit}
is a MCTS that uses UCB1 for action selection and became widely popular in adversarial games.
\citet{schulte2014balancing} proposed several variants of UCT including GreedyUCT (\guct), UCT*, and GreedyUCT* (\guct*).
We often abbreviate a set of algorithms to save space,
e.g., [G]UCT[*] denotes $\braces{\uct, \guct, \uct*, \guct*}$.
In this paper,
we mainly discuss GUCT[*]
due to our focus on the agile satisficing setting that does not prioritize minimization of solution cost.

\subsection{Base MCTS for Graph Search}
\label{sec:mcts-detail}

\refalgo{alg:mcts} shows the pseudocode of MCTS adjusted for graph search, taken from \citep{schulte2014balancing}.
Aside from what was described from the main section, it has a node-locking mechanism
that avoids redundant effort.

Following THTS, our MCTS has a hash table that implements a \emph{CLOSE} list and a \emph{Transposition Table} (TT).
A CLOSE list stores the generated states and avoids instantiating nodes with duplicate states.
A TT stores various information about the states such as the parent information and the action used at the parent.
The close list is implemented by a lock mechanism.

Since an efficient graph search algorithm must avoid visiting the same state multiple times,
MCTS for graph search marks certain nodes as \emph{locked}, and excludes them from the selection candidates.
A node is locked
either (1) when a node is a dead-end that will never reach a goal
(detected by having no applicable actions, by a heuristic function, or other facilities),
(2) when there is a node with the same state in the search tree with a smaller g-value, 
(3) when all of its children are locked,
or
(4) when a node is a goal
(relevant in an anytime iterated search setting \citep{richter2010joy,richter2011lama}, but not in this paper).
Thus, in the expansion step,
when a generated node $n$ has the same state as a node $n'$ already in the search tree,
MCTS
discards $n$ if $g(n)>g(n')$,
else moves the subtree of $n'$ to $n$ and marks $n'$ as locked.
It also implicitly detects a cycle, as this is identical to the duplicate detection in Dijkstra/\astar/GBFS.

The queueing step backpropagates necessary information from the leaf to the root.
Efficient backpropagation uses a priority queue ordered by descending $g$-value.
The queue is initialized with the expanded node $p$; each newly generated node $n$ that is not discarded is inserted into the queue, and if a node $n'$ for the same state was already present in the tree it is also inserted into the queue.
In each backpropagation iteration,
(1) the enqueued node with the highest $g$-value is popped,
(2) its information is updated by aggregating its children's information
(including the lock status),
(3) and its parent is queued.

\begin{algorithm}[tbh]
 \begin{algorithmic}
  \WHILE{True}
  \STATE Parent $p\gets r$
  \WHILE[Selection]{not leaf $p$}
  \STATE $p\gets \argmin_{n\in S(p)} f(n)$
  \ENDWHILE
  \STATE $Q \gets \braces{p}$
  \FOR[Expansion]{$n\in S(p)$}
  \RETURN $n$ \textbf{if} $n$ is goal. \COMMENT{Early goal detection}
  \IF{$\exists n'$ already in tree with same state $s_{n'}=s_n$}
  \IF{$g(n)> g(n')$}\STATE \textbf{continue}
  \ENDIF
  \STATE Lock $n'$, $S(n)\gets S(n')$, $Q \gets Q\cup \braces{n,n'}$
  \ELSE
  \STATE Compute $h(s_{n})$ \COMMENT{Evaluation}
  \STATE $Q \gets Q\cup \braces{n}$
  \ENDIF
  \ENDFOR
  \WHILE[Backpropagation]{$n\gets Q.\function{popmax}()$}
  \STATE Update $n$'s statistics and lock status
  \STATE $Q \gets Q \cup \braces{\text{$n$'s parent}}$
  \ENDWHILE
  \ENDWHILE
 \end{algorithmic}
 \caption{
 High-level general MCTS.
 \textbf{Input}:
 Root node $r$,
 successor function $S$,
 NEC $f$,
 heuristic function $h$,
 priority queue $Q$ sorted by $g$.
 Initialize $\forall n; g(n)\gets \infty$.
}
 \label{alg:mcts}
\end{algorithm}

\section{Existing MCTS-based Classical Planning}
\label{sec:issues}

We revisit GBFS implemented as THTS/MCTS from a MAB perspective.
Let
$S(n)$ be the set of successors of a node $n$,
$L(n)$ be the set of leaf nodes in the subtree under $n$,
and the NECs of \gbfs as $f_{\gbfs}(n)=h_{\gbfs}(n)$.
We expand the definition of the backup functions presented by \citet{keller2013trial}
recursively down to the leaves,
assuming $h_{\gbfs}(n)=h(s_{n})$ if $n$ is a leaf where $h$ is a heuristic.
\begin{align*}
 h_{\gbfs}(n)
 &\textstyle= \min_{n'\in S(n)} [h_{\gbfs}(n')]\\
 &\textstyle= \min_{n'\in S(n)} [\min_{n''\in S(n')} [h_{\gbfs}(n'')]]\\
 =\ldots
 &\textstyle= \min_{n'\in L(n)} [h(s_{n'})].
\end{align*}
\citet{keller2013trial} called the $\min$ operator a \emph{Full-Bellman} backup
and compared it with \emph{Monte-Carlo} backup in GUCT that uses the average,
as expanded to the leaves below as well:
\begin{align*}
 h_{\guct}(n)
 =& \textstyle \frac{1}{|L(n)|}\sum_{n'\in S(n)} |L(n')| h_{\guct}(n') & \\
 =  \textstyle \frac{1}{|L(n)|}\sum_{n'\in S(n)} &\textstyle\frac{\cancel{|L(n')|}}{\cancel{|L(n')|}}\sum_{n''\in S(n')} |L(n'')| h_{\guct}(n'') & \\
 =\ldots
 =& \textstyle \frac{1}{|L(n)|}\sum_{n'\in L(n)} h(s_{n'}). &
\end{align*}
To search, \guct subtracts an exploration term from $h_{\guct}(n)$ based on LCB1,
where $p$ is a parent of $n$.
$|L(p)|$ and $|L(n)|$ respectively correspond to $T$ and $t_i$ in \refeq{eq:ucb1}.
\begin{align*}
 f_{\guct}(n)
 =h_{\guct}(n) -c\sqrt{{(2\log |L(p)|)}/{|L(n)|}} &
\end{align*}

While \citet{keller2013trial} managed to generalize various algorithms
focusing on the procedural aspects (e.g., recursive backup from the children),
we focus on its mathematical meaning.
One key observation missing in \citet{keller2013trial} and is made clear by these expansions
is that
the set $L(n)$ of $n$'s leaves is a \emph{dataset},
the heuristic $h(n')$ at each leaf $n'$ is a \emph{reward sample},
and the NECs estimate its \emph{statistic} such as the mean and the minimum.
(The minimum is known as an \emph{order statistic};
other order statistics include the top-$k$ element, the $q$-quantile, and the median = $0.5$-quantile.)
Backpropagation from the expanded leaves to the root one step at a time
is merely an efficient implementation detail that avoids computing the statistic (min,max,mean) over all leaves every time.
Understanding each $h(n)$
is a sample of a random variable representing a reward for MABs,
we can focus on
the theoretical efficiency guarantees
and see how existing MCTS/THTS for classical planning fail to leverage it.

First,
UCB1 assumes that all reward random variables, each associated with an arm, have a \emph{shared, known bounds},
where each arm corresponds to each successor node during action selections.
Heuristic values in classical planning lack such \emph{a priori} known bounds,
unlike adversarial games whose rewards are either +1/0 or +1/-1 representing a win/loss.
Also,
usually the range of heuristic values in each subtree of the search tree
substantially differ from each other.

Although \citet{schulte2014balancing} claimed to have addressed this issue by modifying the UCB1,
but their modification does not fully address the issue.
Let us call their variant \emph{\guct-01}.
It normalizes the first term of the NEC to $[0,1]$
by taking the minimum and maximum among $n$'s siblings sharing the parent $p$.
Given
 $M=\max_{n'\in S(p)} h_{\guct}(n')$,
 $m=\min_{n'\in S(p)} h_{\guct}(n')$, and a hyperparameter $c$,
\guct-01 modifies $f_{\text{\guct}}$ into $f_{\guct\text{-01}}$ (\refeq{eq:min-max-normalization}).
\begin{align}
 \textstyle
 f_{\guct\text{-01}}(n)
 &\textstyle=\frac{h_{\guct}(n)- m}{M-m} -c\sqrt{\frac{2\log |L(p)|}{|L(n)|}}
 \label{eq:min-max-normalization}
\end{align}
However, the node ordering by the \guct-01's NEC is same when all arms are shifted and scaled by the same amount,
thus \guct-01 is identical to the standard UCB1 with a reward range $[0,c(M-m)]$ (\refeq{eq:min-max-normalization-alt});
we additionally note that this version avoids a division-by-zero issue for $M-m=0$.
\begin{align}
 &\hspace{-2em}m+(M-m)f_{\guct\text{-01}}(n)\notag\\
 &\textstyle= h_{\guct}(n) - c(M-m)\sqrt{\frac{2\log |L(p)|}{|L(n)|}}
 \label{eq:min-max-normalization-alt}
\end{align}
Here are two issues of \guct-01:
First, \guct-01 does not address the fact that different subtrees have different ranges of heuristic values:
When selecting an action, it assumes that all children have the same reward range $[0,c(M-m)]$.
Although $M-m$ differs among parents,
and thus it adjusts its exploration rate in each action selection at a different depth of the tree,
it does not do so for each child,
thus it is depth-aware but not breadth-aware.
Second, we expect \guct-01 to explore excessively,
because the range $[0,c(M-n)]$ obtained from the data of the entire subtree of the parent
is always broader than that of each child,
since the parent's data is a union of those from all children.

Further,
in an attempt to improve the performance of [G]\uct,
\citet{schulte2014balancing} noted that using the average is ``rather odd'' for planning,
and proposed UCT* and GreedyUCT* (GUCT*)
which combines Full-Bellman backup with LCB1
without statistical justification.

Finally,
these variants failed to improve over traditional algorithms (e.g., GBFS)
unless combined with various other enhancements
such as deferred heuristic evaluation (DE) and preferred operators (PO).
The theoretical characteristics of these enhancements are not well understood,
rendering their use ad hoc and the reason for \guct-01's performance inconclusive,
and motivating a better theoretical analysis.

\section{Bandit for Unbounded Distributions}
\label{sec:gaussian-bandit}

To handle reward distributions with unknown supports that differ across arms,
we need a MAB that assumes an unbounded reward distribution spanning the real numbers.
We use the Gaussian distribution here, although future work may consider other distributions.
Formally,
we assume
each arm $i$ has a reward distribution $\N(\mu_i,\sigma^2_i)$ for some unknown $\mu_i,\sigma^2_i$.
As $\sigma^2_i$ differs across $i$, the reward uncertainty differs across the arms.
By contrast, the reward uncertainty of each arm in UCB1 is expressed by the range $[0,c]$, which is the same across the arms. We now discuss the shortcomings of MABs from previous work (\refeqs{eq:ucb1-normal}{eq:ucb-v}), and present our new MAB (\refeq{eq:ucb-normal2}).
\begin{align}
 &\text{UCB1-Normal}_i \textstyle= \hat{\mu}_i + \hat{\sigma}_i\sqrt{ \parens{16 \log T}/{t_i} }
 \label{eq:ucb1-normal} \\
 &\text{UCB1-Tuned}_1 \textstyle= \label{eq:ucb-tuned} \\
 &~~~~~~~~~~~\hat{\mu}_i + c\sqrt{ \min(1/4 , \hat{\sigma}^2_i + \sqrt{2\log T / t_i}) \log T/t_i }
 \nonumber \\
 &\text{UCB-V}_i \textstyle= \hat{\mu}_i + \hat{\sigma}_i \sqrt{ (2\log T)/t_i } + (3c \log T)/t_i
 \label{eq:ucb-v} \\
 &\text{UCB1-Normal2}_i \textstyle= \hat{\mu}_i + \hat{\sigma}_i \sqrt{ 2\log T }
 \label{eq:ucb-normal2}
\end{align}

The UCB1-Normal MAB \citep[Theorem 4]{auer2002finite},
which was proposed along with UCB1 (idem, Theorem 1), is designed exactly for this scenario but is still unpopular.
Given $t_i$ \iid samples $r_{i1}\ldots r_{it_i} \sim \N(\mu_i,\sigma^2_i)$ from each arm $i$
where $T=\sum_i t_i$,
it chooses $i$ that maximizes the metric shown in \refeq{eq:ucb1-normal}.
To apply this bandit to MCTS, substitute $T=|L(p)|$ and $t_i=|L(n)|$,
and backpropagate the statistics $\hat{\mu}_i, \hat{\sigma}^2_i$ (see the appendix \cite[\refsec{app:backprop}]{wissow2023scale:full}).
For minimization tasks such as classical planning, use the LCB.
We refer to the \guct variant using UCB1-Normal as \emph{\guct-Normal}.
An advantage of UCB1-Normal is its logarithmic upper bound on regret \citep[Appendix B]{auer2002finite}.
However, it did not perform well in our empirical evaluation,
likely because its proof relies on two conjectures
which are explicitly stated by the authors as not guaranteed to hold.

\begin{theo}[From \citep{auer2002finite}]
 UCB1-Normal has a logarithmic regret-per-arm
 $256\frac{\sigma_i^2 \log T}{\Delta_i^2}+1+\frac{\pi^2}{2}+8\log T$
 \textbf{if},
 for a Student's $t$ RV $X$ with $s$ degrees of freedom (DOF),
 $\forall a \in [0, \sqrt{2(s+1)}]; P(X\geq a)\leq e^{-a^2/4}$,
 and \textbf{if},
 for a $\chi^2$ RV $X$ with $s$ DOF,
 $P(X\geq 4s)\leq e^{-(s+1)/2}$.
\end{theo}

To avoid relying on these two conjectures,
we need an alternate MAB that similarly adjusts the exploration rate based on the variance.
Candidates include
UCB1-Tuned \citep{auer2002finite} in \refeq{eq:ucb-tuned},
and
UCB-V \citep{audibert2009exploration} in \refeq{eq:ucb-v},
but they all have various limitations.
UCB1-Tuned assumes a bounded reward distribution, lacks a regret bound, and is outperformed by UCB-V.
UCB-V improves UCB1-Tuned with a regret proof but it still assumes a bounded reward distribution.

\begin{figure}[tb]
 \centering
 \includegraphics[width=0.9\linewidth]{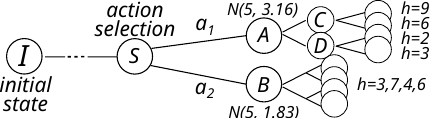}
 \includegraphics[width=0.9\linewidth]{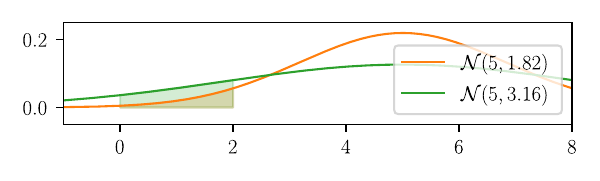}
 \caption{
 Search behavior.
 Larger $\sigma$ assigns more probability on the lower $h$ values we get in the next expansion.
 }
 \label{fig:exploration}
\end{figure}

We present UCB1-Normal2 (\refeq{eq:ucb-normal2})
and analyze its regret bound (\emph{which is one of our main contributions}).
To understand its behavior,
see \refig{fig:exploration} which shows a MCTS selecting an action at a state $S$
(which is equivalent to selecting a node from the open list, see \refsec{sec:mcts}).
$S$ has two successors $A$ and $B$.
The subtrees of $A$ and $B$ contain $h$-values $\{9,2,6,3\}$ and $\{3,7,4,6\}$,
therefore their backpropagated statistics are $\mathcal{N}(\mu,\sigma)=\mathcal{N}(5,3.16)$ and $\mathcal{N}(5,1.83)$.
$A$ has a larger $\sigma$, thus
the algorithm expects a higher chance of finding a lower $h$-value under $A$, as shown in the green plot.
Although there is also a higher chance of finding a bad node (high $h$) under $A$,
it avoids such branches during further recursions (e.g., prefer $D$ over $C$).
In other words,
when the means are equal, it avoids low variances
which is likely to contain mediocre heuristic values $h\approx 5$ all the time.
This mechanism generalizes the concept of escaping heuristic plateaus during search \citep{Coles07}:
The low $\sigma$ of $B$ indicates that its subtree contains a set of nodes with different but similar values.
A heuristic plateau is a special case of $\sigma=0$.
In contrast, GUCT does not use $\sigma$ and prefers $A$ and $B$ equally
(both nodes have LCB $=\mu-c\sqrt{\frac{2\log T}{n_i}}=5-c\sqrt{\frac{2\log 8}{4}}$).

\begin{theo}[Main Result]
 Let
 $\alpha\in [0,1]$ be an unknown problem-dependent constant
 and $\chi^2_{1-\alpha,n}$ be the critical value for the tail probability of a $\chi^2$ distribution with significance $\alpha$ and DOF $n$
 that satisfies $P({t_i\hat{\sigma}_i^2}/{\sigma_i^2} < \chi^2_{1-\alpha,t_i})=\alpha$.
 UCB1-Normal2 has a following worst-case polynomial, best-case constant regret-per-arm
 where
 $C$ is a finite constant if each arm is pulled at least
 $M=\inf \{n | 8<\chi^2_{1-\alpha,n}\}$ times.
 \begin{align*}
  \frac{-4(\log \alpha)\sigma_i^2\log T}{\Delta_i^2} +& 1+2C+\frac{(1-\alpha) T(T+1)(2T+1)}{3}
  \\
  &\to[\alpha\to 1] 1+2C
 \end{align*}
\end{theo}
\begin{proof}
 \emph{(Sketch of the appendix \cite[\refsec{app:ucb-normal2-proof}]{wissow2023scale:full}.)}
 We use Hoeffding's inequality for sub-Gaussian distributions as
 Gaussian distributions belong to sub-Gaussian distributions.
 Unlike in UCB1 where the rewards have a fixed known support $[0,c]$,
 we do not know the true reward variance $\sigma_i^2$.
 Therefore,
 we use the mathematical truth \emph{(not a conjecture)} that,
 when $r_{i1}\ldots r_{it_i} \sim \N(\mu_i,\sigma^2_i)$
 and $\hat{\sigma_i}^2= \frac{1}{t_i}\sum_{j=1}^{t_i} r^2_{ij} - (\frac{1}{t_i}\sum_{j=1}^{t_i} r_{ij})^2$,
 then ${t_i\hat{\sigma_i}^2}/{\sigma_i^2}$ follows a $\chi^2$ distribution and
 $P({t_i\hat{\sigma_i}^2}/{\sigma_i^2} < \chi^2_{1-\alpha,t_i})=\alpha$ for some $\alpha$.
 We use union-bound to address the correlation and further upper-bound the tail probability.
 We also use $\chi^2_{1-\alpha,t_i}\geq \chi^2_{1-\alpha,2}=-2\log \alpha$ for $t_i\geq 2$.
 The resulting upper bound contains an infinite series $C$.
 Its convergence condition dictates the minimum pulls $M$ that must be performed initially.
\end{proof}

Polynomial regrets are generally worse than logarithmic regrets of UCB1-Normal.
However,
UCB1-Normal relies on unproven conjectures and our experimental results shows 
UCB1-Normal is outperformed by UCB1-Normal2, suggesting that these conjectures do not hold.
Our regret bound also improves over that of UCB1-Normal if $T$ is small and $\alpha\approx 1$ ($\log \alpha\approx 0$ therefore $1-\alpha\approx 0$).
$\alpha$ represents the accuracy of the sample variance $\hat{\sigma}^2$ toward the true variance $\sigma^2$.
In deterministic, discrete, finite state-space search problems like classical planning,
$\alpha$ tends to be close to (or sometimes even match) 1
because $\sigma=\hat{\sigma}$ is achievable.
Several factors of classical planning contribute to this.
Heuristic functions in classical planning are deterministic,
unlike rollout-based heuristics in adversarial games.
This means
$\sigma=\hat{\sigma}=0$ when a subtree is linear due to the graph shape.
Also, $\sigma=\hat{\sigma}$ when all reachable states from a node are exhaustively enumerated in its subtree.
In statistical terms,
this is because draws from heuristic samples are performed without replacements due to duplication checking in search algorithms.

Unlike UCB1-Normal, which pulls arms uniformly until all arms satisfy $t_i\geq \lfloor 8\log T\rfloor$,
UCB1-Normal2 does not need such initialization pulls
because every node is evaluated once and its heuristic value is used as a single sample.
This means we assume $M=1$, thus $\alpha > \erf(2)>0.995$ because
$8<\chi^2_{1-\alpha,1} \iff 1-\alpha < \frac{\gamma(\frac{1}{2}, \frac{8}{2})}{\Gamma(\frac{1}{2})}=1-\erf(2)$. 
In classical planning,
$\alpha>0.995$ is more realistic than the conjectures used by UCB1-Normal.

Another explanation for the failure of UCB1-Normal is that
its exploration term is made too confident / too small by ${1}/{\sqrt{t_i}}$
because it was derived with more assumptions (the $\chi^2$ conjecture).

During the discussion,
yet another potential explanation related to \emph{the aleatoric and epistemic uncertainty}
was suggested by one of the reviewers.
The rewards in the typical bandit problems are ``truly'' stochastic,
i.e., each arm always samples a different reward from the unknown fixed distribution,
thus the uncertainty is aleatoric / an objective truth.
However all rewards in classical planning are deterministic,
and the uncertainty comes purely from the sampling (search behavior in the subtree),
thus the uncertainty is epistemic / subjective to the agent.
Under this interpretation, the standard regret analysis above (yielding polynomial regret)
may tell little about the actual performance of the algorithms,
potentially suggesting a new challenge for the bandit community.

\section{Experimental Evaluation}
\label{sec:experiments}

We evaluated the algorithms over
a subset of the International Planning Competition benchmark domains,\footnote{\url{github.com/aibasel/downward-benchmarks}}
selected for compatibility with the set of PDDL extensions supported by Pyperplan \citep{pyperplan}.
We maintain the superset of the results of the experiments under a 10,000 node evaluation limit,
a 4,000 node expansion limit, and a 300 second runtime limit, and then
count the number of instances solved under each limit.
We mainly focus on the node evaluations
because heuristic computation is the main bottleneck in classical planning.
See the appendix \cite[\refigs{fig:evaluation-histogram}{fig:elapsed-histogram}]{wissow2023scale:full} for the results
controlled by expansions and the runtime.
Another reason for this focus is the fact that
we used a python-based implementation (Pyperplan) for convenient prototyping.
It is slower than C++-based \lsota systems (e.g. Fast Downward \citep{Helmert2006}),
but our focus on evaluations makes this irrelevant and also improves reproducibility by avoiding the effect of hardware differences and low-level implementation details.

\begin{table*}[tb]
\caption{
 The number of problem instances solved with less than 10,000 node evaluations;
 best configurations in \textbf{bold} for each heuristic;
 each number represents an average over 5 trials.
 We show results for both $c=1.0$ and $c=0.5$ (``best parameter'' according to \citet{schulte2014balancing})
 when the algorithm requires one.
 Algorithms in the bottom half have no hyperparameter.
 $\dagger$: Data missing due to the lack of support of PO for GBFS in Pyperplan.
 $\ddagger$: Data missing because DE in Fast Downward measures node evaluations differently.
 }
 \centering
 {\small PO:Preferred Operators, DE:Deferred Evaluation.}
\begin{adjustbox}{width=\linewidth}
  \begin{tabular}{l*{7}{cc}}
  \toprule
  \multicolumn{1}{r}{$h=$}
  & \multicolumn{2}{c}{$\ff$}
  & \multicolumn{2}{c}{$\ad$}
  & \multicolumn{2}{c}{$\hmax$}
  & \multicolumn{2}{c}{$\gc$}
  & \multicolumn{2}{c}{$\ff$+PO}
  & \multicolumn{2}{c}{$\ff$+DE}
  & \multicolumn{2}{c}{$\ff$+DE+PO}
  \\
  \cmidrule(r){2-3}
  \cmidrule(r){4-5}
  \cmidrule(r){6-7}
  \cmidrule(r){8-9}
  \cmidrule(r){10-11}
  \cmidrule(r){12-13}
  \cmidrule(r){14-15}
  \multicolumn{1}{r}{$c=$}
 & 0.5 & 1 & 0.5 & 1 & 0.5 & 1 & 0.5 & 1 & 0.5 & 1 & 0.5 & 1 & 0.5 & 1 \\
\midrule
GUCT & 442.8 & 412.0 & 435.8 & 397.8 & 237.0 & 228.4 & 306.6 & 285.2 & 484.6 & 454.0 & 455.8 & 389.2 & 497.4 & 439.4 \\ 
* & 542.0 & 458.6 & 529.2 & 480.8 & 248.4 & 242.2 & 317.8 & 310.4 & 591.6 & 495.8 & 480.2 & 423.6 & 527.4 & 471.0 \\ 
-01 & 399.8 & 368.0 & 375.4 & 328.8 & 256.8 & 237.4 & 318.4 & 302.4 & 441.4 & 408.4 & 387.6 & 361.2 & 445.0 & 422.6 \\ 
*-01 & 425.6 & 388.0 & 404.8 & 364.4 & 246.8 & 233.4 & 318.0 & 297.6 & 470.6 & 420.6 & 409.4 & 378.2 & 466.8 & 438.8 \\ 
-V & 361.2 & 317.4 & 354.0 & 310.6 & 226.2 & 208.6 & 278.4 & 255.4 & 427.0 & 389.6 & 370.8 & 344.6 & 431.2 & 421.0 \\ 
\midrule
-Normal & - & 283.4 & - & 265.0 & - & 212.0 & - & 233.4 & - & 372.4 & - & 289.0 & - & 381.6 \\ 
*-Normal & - & 318.8 & - & 300.0 & - & 215.2 & - & 246.2 & - & 378.1 & - & 304.4 & - & 386.7 \\ 
-Normal2 & - & \textbf{581.8} & - & 535.8 & - & \textbf{316.6} & - & \textbf{379.0} & - & \textbf{621.0} & - & \textbf{518.0} & - & \textbf{578.0} \\ 
*-Normal2 & - & 567.2 & - & 533.8 & - & 263.0 & - & 341.0 & - & 618.0 & - & 511.4 & - & 567.8 \\ 
TTTS-Normal & - & 181.0 & - & 180.0 & - & 171.4 & - & 170.8 & - & 151.0 & - & 180.6 & - & 150.6 \\ 
TTTS-Normal* & - & 189.4 & - & 186.4 & - & 177.4 & - & 174.4 & - & 159.4 & - & 185.8 & - & 155.8 \\ 
\midrule
\multicolumn{2}{l}{GBFS(Pyperplan/FastDownward)} & 538/539 & - & 518/517 & - & 224/226 & - & 354/349 & - & $\dagger$/539 & - & 489/$\ddagger$ & - & $\dagger$/$\ddagger$ \\ 
\multicolumn{2}{l}{W\astar$(w=5)$ (FastDownward)}&528&-& 522&-& 211&-& 319&-& 528&-& $\ddagger$&-& $\ddagger$\\
\multicolumn{2}{l}{Softmin-Type(h) (FastDownward)} & 576.0 & - & \textbf{542.6} & - & 297.2 & - & 357.6 & - & 575.8 & - & $\ddagger$ & - & $\ddagger$ \\ 
  \bottomrule
 \end{tabular}
\end{adjustbox}
 \label{tbl:main-table}
\end{table*}

In order to limit the length of the experiment,
we also removed the problem instances
which Pyperplan took more than 5 minutes and 2GB memory to parse and instantiate the input file.
The instantiation limit removed
47 instances from freecell,
4 from logistics98,
2 from openstacks, and
24 from pipesworld-tankage.
This resulted in 772 problem instances across 24 domains in total.
We evaluated various algorithms with $\ff$, $\ad$, $\hmax$, and $\gc$ (goal count) heuristics \citep{FikesHN72},
and our analysis focuses on $\ff$.
We included $\gc$ because it can be used in environments without domain descriptions,
e.g., in the planning-based approach \citep{lipovetzky2015classical} to the Atari environment \citep{bellemare2015arcade}.
We ran each configuration with 5 random seeds and report the average number of problem instances solved.
To see the spread due to the seeds, see the cumulative histogram plots in the appendix \cite[\refigs{fig:evaluation-histogram}{fig:elapsed-histogram}]{wissow2023scale:full}.

We evaluated the following algorithms:
\textbf{\gbfs} is \gbfs implemented in Pyperplan and FastDownward.
We evaluated both implementations in order to compare the difference.
\textbf{WA*} ($w=5$) based on FastDownward is added because
it outperformed GBFS in \citep{schulte2014balancing} in agile setting.
\textbf{\guct} is a \guct based on the original UCB1.
\textbf{\guct-01} is \guct with ad hoc $[0,1]$ normalization of the mean \citep{schulte2014balancing}.
\textbf{\guct-Normal/-Normal2/-V} are \guct variants using UCB1-Normal/UCB1-Normal2/UCB-V respectively.
The * variants \textbf{\guct*/-01/-Normal/-Normal2} are using full-bellman backup.
For \guct and \guct-01, we evaluated the hyperparameter $c$ with the standard value $c=1.0$ and $c=0.5$.
The choice of the latter is due to \citet{schulte2014balancing},
who claimed that \guct[*]-01 performed the best when $0.6<C=c\sqrt{2}<0.9$, i.e., $0.4<c<0.63$.
Our aim of testing these hyperparameters is
to compare them against automatic exploration rate adjustments performed by UCB1-Normal[2].
Other algorithms are explained later.

\citet{schulte2014balancing} previously reported that
two ad hoc enhancements to GBFS, PO and DE, also improve the performance of \guct[*]-01.
We evaluated our equivalent reimplementation.
We did not evaluate PO with heuristics other than $\ff$, which are not supported by Pyperplan.

In all comparisons between \guct-Normal2 and other algorithms based on FF heuristics below,
we performed Welch's unequal variances $t$-test 
on the coverage scores with 5 different random seeds, and confirmed $p<0.001$ for all comparisons.
Note that all algorithms evaluated in this paper are deterministic up to tie-breaking, including MCTS variants:
With the same set of leafs, all NECs are deterministic, and thus the action selection is deterministic.

\paragraph{Detailed Ablation}
We first reproduced \citep{schulte2014balancing} and provide its more detailed ablation.
\reftbl{tbl:main-table} shows that \guct[*]-01 is indeed significantly outperformed by
the baseline algorithm \gbfs,
indicating that UCB1-based exploration is not beneficial for planning.
Although this result disagrees with the final conclusion of their paper,
their conclusion relied on incorporating the DE and PO enhancements,
and these confounding factors impede conclusive analysis.

We also tested \guct[*], which lacks 
the mean normalization (\refeq{eq:min-max-normalization}) of \guct[*]-01,
which was not previously evaluated.
\guct[*]-01 performs significantly worse,
indicating that its normalization
not only fails to address the unknown and different supports,
but also harms the performance by excessive exploration,
as predicted by our analysis in \refsec{sec:issues}.

\paragraph{\guct-Normal2}
We then compared various algorithms.
\guct-Normal2 outperformed \gbfs,
\guct/-01/-Normal/-V, and their * variants.
The dominance against \guct-Normal
supports our analysis that in classical planning
$\hat{\sigma}^2\approx\sigma^2$, thus $P(t_i\hat{\sigma}^2/\sigma^2<\chi^2_{1-\alpha,t_i})=\alpha\approx 1$,
overcoming the asymptotic deficit
(the polynomial regret in \guct-Normal2 vs.\ the logarithmic regret of \guct-Normal).
In other words, the logarithmic regret of UCB1-Normal does not hold in classical planning
because the $\chi^2$ conjectures tend to be violated.

While the * variants can be significantly better than the non-* variants at times,
this trend was opposite in algorithms that perform better,  e.g., \guct*-Normal2 tend to be worse than \guct-Normal2.
This supports our claim that
Full-Bellman backup proposed by \citep{schulte2014balancing} is theoretically unfounded and thus
does not consistently improve search algorithms.
Further theoretical investigation of Full-Bellman backup is an important avenue of future work.

The table also compares \guct[*]-Normal[2], which do not require any hyperparameter,
against \guct[*][/-01/-V] with different $c$ values.
Although $c=0.5$ improves the performance of \guct[*]-01 as reported by \citet{schulte2014balancing},
it did not improve enough to catch up with the adaptive exploration rate of \guct[*]-Normal2.
We also tested $c\in\braces{0.1,0.3,3.0}$ (see the appendix \cite[\refsec{sec:different-c}]{wissow2023scale:full}).
Results indicated that $c=0.1$ tends to be better, but it still did not outperform \guct-Normal2
(e.g., the best coverage among \guct[*][-01] with \ff/$c=0.1$ was 561.8 by \guct*,
compared to 581.8 by \guct-Normal2.)
This is not surprising, as the limit of $c\to 0$ for \guct* is GBFS, which also performs well (538).

\paragraph{Simple Regret}
\emph{Cumulative regret} (CR, \refeq{def:cr}) bandits
maximize the total rewards, \emph{including those obtained during the experimentation},
while 
\emph{simple regret} (SR) bandits / \emph{best-arm identification} algorithms \citep{audibert2010best,karnin2013almost}
maximize the expected rewards of the \emph{incumbent best arm} 
that is maintained during the experimentation and is returned at the end.
SR MABs can explore more aggressively than CR MABs
because the cost of the experimentation
 is free for SR.
\citet{feldman2014simple} showed that SR MABs are superior in \emph{online} MDPs where
the incumbent corresponds to the action to take next.
The concept of incumbent also exists in anytime search \citep{richter2010joy},
e.g., in LAMA \citep{richter2011lama}, suggesting an interesting future direction.
However, the incumbent does not exist in agile search, or in the first iteration of anytime search,
so SR MABs are conceptually mismatched with offline agile search.
\reftbl{tbl:main-table} shows the performance of MCTS combined with a \lsota SR MAB  \citep[TTTS]{russo2020simple}.
TTTS-Normal/* respectively uses the Monte-Carlo/Full-Bellman backup.
As expected, they are vastly outperformed by other algorithms.
In online planning and acting,
the justification for SR is that search is not a commitment, only the first action of the policy/plan is.
\emph{However, in agile planning, search is indeed a commitment to the computational resources (time, memory) which we minimize,}
thus it justifies the CR objective.

\paragraph{Preferred Operators}
In addition to the heuristic value of a state,
some heuristic functions are able to return a list of actions called 
``helpful actions'' \citep{hoffmann01} or ``preferred operators'' \citep{richter2009preferred}.
We reimplemented \citet{schulte2014balancing}'s strategy which
limits the action selection to the preferred operators
and falling back to the normal behavior if there are none.
\reftbl{tbl:main-table} shows that it also improves \guct[*][-Normal2],
consistent with the previous report on \guct[*]-01.

\paragraph{Deferred Evaluation}
\reftbl{tbl:main-table} shows the effect of deferred heuristic evaluation (DE) on search algorithms.
\emph{In this experiment,}
DE should perform worse than eager evaluations
because DE trades the number of calls to heuristics
with the number of nodes inserted to the tree, which is limited to 10,000.
When CPU time is the limiting resource, DE usually solves more instances,
assuming the implementation is optimized for speed (e.g., using C++).
However, our Python implementation (typically 100--1,000 times slower than C++)
is not able to measure this effect
because
this low-level bottleneck could hide the effect of speed improvements.
What we could learn from this experiment is therefore 
whether DE+PO is better than DE, and
if \guct[*]-Normal2+DE continues to dominate other algorithms with DE.
\reftbl{tbl:main-table} answers both questions positvely:
DE+PO tends to perform better than DE alone, and
\guct[*]-Normal2 is still superior to other algorithms with DE and DE+PO.

\paragraph{Diversified Search}
We evaluated Softmin-Type(h) \citep{kuroiwa2022biased},
a recent \lsota diversified search algorithm for classical planning.
We used the original C++ implementation based on Fast Downward.
We excluded diversification methods that use state information, such as BFWS \citep{lipovetzky2017bwfs},
as they are orthogonal concepts.
\reftbl{tbl:main-table} shows that UCB1-Normal2 outperforms Softmin-Type(h) with $\ff,\hmax,\gc,\ff\textrm{+PO}$.
See the domain-wise comparisons in the appendix \cite[\refsec{sec:domain-wise}]{wissow2023scale:full}.

\begin{figure}[tb]
 \centering
 \includegraphics[width=0.32\linewidth]{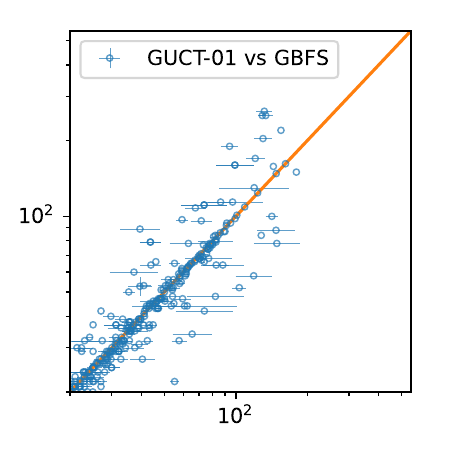}
 \includegraphics[width=0.32\linewidth]{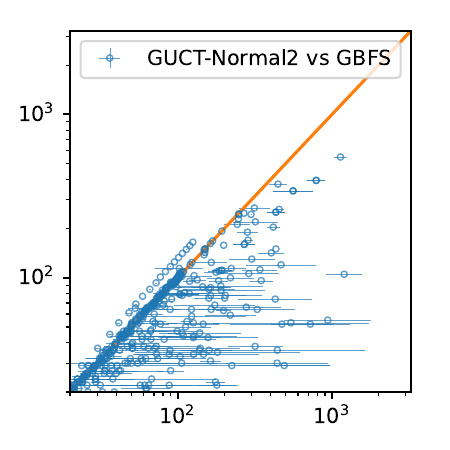}
 \includegraphics[width=0.32\linewidth]{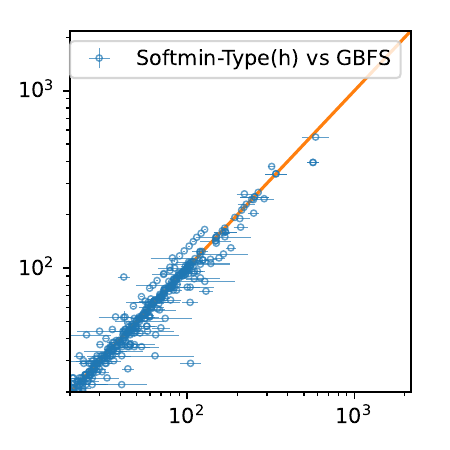}
 \caption{
 Comparing solution length of \guct[-01/Normal2] and Softmin-Type(h) ($x$-axis)
 against \gbfs ($y$-axis) using $\ff$.
 }
 \label{fig:solution-length}
\end{figure}

\paragraph{Solution Quality}
\refig{fig:solution-length} shows that
\guct-Normal2 and Softmin-Type(h) return longer solutions than \gbfs does.
\guct-01 
finds solutions with highly varying length, but overall they are not consistently longer or shorter than \gbfs.
See the appendix \cite[\refigs{fig:solution-length-hff}{fig:solution-length-gc}]{wissow2023scale:full} for more plots.
For agile search, we believe
that a successful exploration must sacrifice the solution quality for faster search.

\paragraph{Runtime Comparison}

To assess the impact of the runtime overhead required by \guct-Normal2 to maintain MCTS search tree,
we reimplemented \guct/-Normal/2 on Fast Downward and evaluated them on IPC 2018 satisficing instances.
\reftbl{tbl:ipc18-sat} shows that \guct-Normal2 outperforms other algorithms.

\begin{table}[htbp]
\caption{IPC 2018 results (average number of instances solved over 3 seeds)
using $\ff$ under 5 min time limit and 8GB memory limit.
For caldera and organic-synthesis,
we used their action-splitting variants \cite{areces2014optimizing}
provided by the organizers. ``Softmin'' stands for Softmin-Type(h).
Best results are highlighted in bold.}
\begin{tabular}{r|rrr|rrr}
\toprule
domain    & GBFS          & WA*           & Softmin       & GUCT         & Normal        & Normal2       \\ \midrule
agricola  & 9.0           & 4.0           & 9.0           & 8.0          & 1.0           & \textbf{9.7}  \\ 
caldera   & 4.0           & 2.0           & \textbf{7.3}  & 6.0          & 6.7           & 6.7           \\ 
data-net  & 4.0           & 5.0           & 9.0           & 4.0          & 2.0           & \textbf{9.7}  \\ 
flashfill & 9.0           & 8.0           & \textbf{9.0}  & 1.0          & 0.0           & 6.7           \\ 
nurikabe  & 7.0           & 8.0           & 7.0           & 8.0          & 8.0           & \textbf{8.3}  \\ 
org-syn   & 9.0           & {10.0}        & 9.3           & {10.0}       & \textbf{10.3} & {9.7}         \\ 
settlers  & 0.0           & 4.0           & {5.3}         & \textbf{6.3} & 5.3           & 2.3           \\ 
snake     & 5.0           & 3.0           & 5.0           & 3.3          & 3.0           & \textbf{16.7} \\ 
spider    & 8.0           & \textbf{11.0} & 8.7           & 8.7          & 9.0           & {9.3}         \\ 
termes    & \textbf{12.0} & 4.0           & \textbf{12.0} & 10.0         & 9.7           & 6.0           \\ \midrule
total     & {67.0}        & {59.0}        & {81.7}        & {65.3}       & {55.0}        & \textbf{85.0} \\
\bottomrule
\end{tabular}
\label{tbl:ipc18-sat}
\end{table}

\section{Related Work}
\label{sec:relawork}

The idea of using variances to guide the search has been proposed as early as
Crazy Stone for computer Go \citep{coulom2006efficient}.
However, due to its focus on adversarial games,
MCTS literature typically assumes a bounded reward setting (e.g., 0/1, -1/+1),
making applications of UCB1-Normal scarce
(e.g., Google Scholar returns 5900 vs.\ 60 for keyword ``UCB1'' and ``UCB1-Normal'', respectively)
except a few model-selection applications \citep{mcconachie2018estimating}.
While Gaussian Process MAB \citep{srinivas2010gaussian} has been used with MCTS for
sequential decision making in continuous space search and robotics \citep{kim2020monte},
it is significantly different from 
discrete search spaces like in classical planning.

MABs may provide a rigorous theoretical tool to analyze the behavior
of a variety of existing randomized enhancements for agile/satisficing search
that tackle the exploration-exploitation dilemma.
$\epsilon$-greedy GBFS was indeed inspired by MABs \citep[Sec.2]{valenzano2014comparison}.

\guct-Normal2 encourages exploration in nodes further from the goal,
which tend to be close to the initial state.
This behavior is similar to that of Diverse Best First Search \citep{imai2011novel},
which stochastically enters an ``exploration mode''
that expands a node with a smaller $g$ value more often.
This reverse ordering is unique from other diversified search algorithms,
including $\epsilon$-GBFS,
Type-GBFS \citep{xie15understanding}, and Softmin-Type-GBFS \citep{kuroiwa2022biased},
which selects $g$ rather uniformly during the exploration.

Theoretical guarantees of MABs require modifications
in tree-based algorithms (e.g. MCTS) due to non-\iid sampling from the subtrees \citep{coquelin2007bandit,munos2014bandits}.
Incorporating the methods developed in the MAB community to counter this bias in the subtree samples is
an important direction for future work.

MDP and Reinforcement Learning literature often use discounting
to avoid the issue of divergent cumulative reward:
when the upper bound of step-wise reward is known to be $R$, then
the maximum cumulative reward goes to $\infty$ with infinite horizon,
while the discounting with $\gamma$ makes it below $\frac{R}{1-\gamma}$,
allowing the application of UCB1.
Although it addresses the numerical issue and UCB1's theoretical requirement,
it no longer optimizes the cumulative objective.

\section{Conclusion}

We examined the theoretical assumptions of existing bandit-based exploration mechanisms for classical planning,
and showed that ad hoc design decisions can invalidate theoretical guarantees and harm performance.
We presented \guct-Normal2,
a classical planning algorithm combining MCTS and our Gaussian bandit UCB1-Normal2,
and analyzed it both theoretically and empirically.
Future work includes combinations with other enhancements for agile search
including novelty metric \citep{lipovetzky2017bwfs},
lazy evaluations and preferred operators \citep{RichterH2009},
and iterated anytime search \citep{richter2010joy}.

\section*{Acknowledgments}

This work was supported through DTIC contract FA8075-18-D-0008, Task
Order FA807520F0060, Task 4 - Autonomous Defensive Cyber Operations
(DCO) Research \& Development (R\&D).

\appendix

\fontsize{9.5pt}{10.5pt}
\selectfont

\clearpage
\renewcommand{\thesection}{S\arabic{section}}
\renewcommand{\thetable}{S\arabic{table}}
\renewcommand{\thefigure}{S\arabic{figure}}

\section*{Appendix}

\section{Domain-Independent Heuristics in Classical Planning}
\label{sec:heuristics}

A domain-independent heuristic function $h$ in classical planning is
a function of a state $s$ and the problem $\brackets{P,A,I,G}$,
but the notation $h(s)$ usually omits the latter.
In addition to what we discussed in the main article, this section also uses a notation $h(s,G)$.
It returns an estimate of the cumulative cost from $s$ to one of the goal states (states that satisfy $G$),
typically through a symbolic, non-statistical means including problem relaxation and abstraction.
Notable \lsota functions that appear in this paper includes
$\ff, \hmax, \ad, \gc$ \cite{hoffmann01,bonet2001planning,FikesHN72}.

A significant class of heuristics is called delete relaxation heuristics,
which solve a relaxed problem which does not contain delete effects,
and then returns the cost of the solution of the relaxed problem as an output.
The cost of the optimal solution of a delete relaxed planning problem from a state $s$ is
denoted by $h^+(s)$, but this is too expensive to compute in practice (NP-complete) \cite{bylander1996}.
Therefore, practical heuristics typically try to obtain its further relaxations
that can be computed in polynomial time.

One such admissible heuristic based on delete-relaxation is called
$\hmax$ \cite{bonet2001planning} that is recursively defined as follows:

\begin{align}
 \hmax(s,G) = \max_{p\in G}
 \left\{
  \begin{array}{l}
   0\ \text{if}\ p\in s.\ \text{Otherwise,}\\
   \min_{\braces{a\in A\mid p\in\adde(a)}} \\
    \quad \left[\cost(a)+\ad(s, \pre(a))\right].
  \end{array}
 \right.
\end{align}

Its inadmissible variant is called
additive heuristics $\ad$ \cite{bonet2001planning} that is recursively defined as follows:

\begin{align}
 \ad(s,G) = \sum_{p\in G}
 \left\{
  \begin{array}{l}
   0\ \text{if}\ p\in s.\ \text{Otherwise,}\\
   \min_{\braces{a\in A\mid p\in\adde(a)}} \\
    \quad \left[\cost(a)+\ad(s, \pre(a))\right].
  \end{array}
 \right.
\end{align}

Another inadmissible delete-relaxation heuristics called
$\ff$ \cite{hoffmann01} is defined based on another heuristics $h$, such as $h=\ad$, as a subprocedure.
For each unachieved subgoal $p\in G\setminus s$,
the action $a$ that adds $p$ with the minimal $\left[\cost(a)+h(s, \pre(a))\right]$
is conceptually ``the cheapest action that achieves a subgoal $p$ for the first time under delete relaxation'',
called the \emph{cheapest achiever} / \emph{best supporter} $\text{bs}(p,s,h)$ of $p$.
$\ff$ is defined as the sum of actions in a relaxed plan $\Pi^+$ constructed as follows:

\begin{align}
 \ff(s,G,h) &= \sum_{a\in \Pi^+(s,G,h)} \cost(a)\\
 \Pi^+(s,G,h) &= \bigcup_{p\in G}
 \left\{
  \begin{array}{l}
   \emptyset\ \text{if}\ p\in s.\ \text{Otherwise,}\\
   \braces{a} \cup \Pi^+(s,\pre(a)) \\
   \qquad \text{where}\ a=\text{bs}(p,s,h).
  \end{array}
 \right.\\
 \text{bs}(p,s,h)&=\argmin_{\braces{a\in A\mid p\in \adde(a)}} \left[\cost(a)+h(s, \pre(a))\right].
\end{align}

Goal Count heuristics $\gc$ is a simple heuristic proposed in \cite{FikesHN72}
that counts the number of propositions that are not satisfied yet.
$\brackets{\text{condition}}$ is a cronecker's delta / indicator function that returns 1 when the condition is satisfied.
\begin{align}
 \gc(s,G) &= \sum_{p\in G} \dbrackets{p\not \in s}.
\end{align}

\clearpage
\section{Proof of Bandit Algorithms (Tutorial)}
\label{app:ucb-general-proof}

\emph{This section serves as a tutorial for understanding our main proof in the later sections.}
To help understand the proof of various confidence bounds,
we first describe the general procedure for proving the regret of bandit algorithms,
demonstrate the proof of UCB1 using this scheme,
then finally show the proof of other bandits.

The ingredients for proving an upper/lower confidence bound are as follows:

\begin{itemize}
 \item \textbf{Ingredient 1: The main term and the exploration term.}
       For example, in the standard UCB1 \citep{auer2002finite},
       the \emph{main term} is the empirical mean $\hat{\mu}$
       while the \emph{exploration term} is $c\sqrt{\frac{2\log T}{n_i}}$.
       Their forms are heavily affected by the proof of the upper bound on the regret,
       therefore it is not like you can use an arbitrary exploration term you came up with.
 \item \textbf{Ingredient 2: A specification of reward distributions.}
       For example, in the standard UCB1 \citep{auer2002finite}, one assumes a reward distribution bounded in $[0,c]$.
       Different algorithms assume different reward distributions,
       and in general, more information about the distribution gives a tighter bound (and faster convergence).
       For example, one can assume an unbounded distribution with known variance, etc.
 \item \textbf{Ingredient 3: A concentration inequality.}
       It is also called a tail probability bound.
       For example, in the standard UCB1, one uses Hoeffding's inequality.
       Different algorithms use different inequalities to prove the bound,
       based on what reward distribution it assumes and what main term it uses.
       Examples include the Chernoff bound, Chebishev's inequality, Bernstein's inequality,
       Bennett's inequality, etc.
       Note that the inequality may be two-sided or one-sided.
\end{itemize}

The general procedure for proving the bound is as follows.

\begin{enumerate}
 \item Let the main term be a random variable $\rX_n$
       and the exploration term be $\delta$.
       Then write down the concentration inequality for $\rX_n$ as follows.
       \begin{itemize}
        \item $P(|\rX_n-\E[\rX_n]|\geq \delta)\leq F(\delta).$ (two-sided)
       \end{itemize}
       $F$ is an inequality-specific formula.
       If necessary, simplify the inequality based on the assumptions made in the reward distribution,
       e.g., bounds, mean, variance.
 \item Expand $|\rX_n-\E[\rX_n]|\geq\delta$ into $\delta \geq \rX_n-\E[\rX_n] \geq -\delta$.
 \item Change the notations to model the bandit problem
       because each concentration inequality is a general statement about RVs.
       Before this step, the notation was:
       \begin{itemize}
        \item $n$ (number of samples)
        \item $\rX_n$ is a function of \iid random variables $(\rx_1, \ldots, \rx_n)$
        \item $\E[\rX_n]=\E[\rx_1]=\ldots=\E[\rx_n]$ is assumed.
       \end{itemize}
       For example,
       \begin{itemize}
        \item $\mu_n=\frac{1}{n}\sum_{i=1}^{n}\rx_i$
        \item $\E[\mu_n]=\E[\rx_1]=\ldots=\E[\rx_n]$
       \end{itemize}
       After the change, they correspond to:
       \begin{itemize}
        \item $n_i$ (number of pulls of arm $i$).
        \item $\hat{X}_i$ (empirical value of arm $i$ from $n_i$ pulls),
        \item $X_i$ (true value of arm $i$).
       \end{itemize}
       For example,
       \begin{itemize}
        \item $\hat{\mu}_i$ (sample mean of arm $i$ from $n_i$ pulls),
        \item $\mu_i$ (true mean of arm $i$),
       \end{itemize}
 \item Let $i$ be a suboptimal arm, $*$ be an optimal arm,
       $\mathrm{UCB}_i=\hat{\mu}_i+\delta$, and
       $\mathrm{LCB}_i=\hat{\mu}_i-\delta$.
       Derive the relationship between $\delta$ and the gap $\Delta_i=\mu_i-\mu_*$
       so that the following conditions for the best arm holds:
       \begin{itemize}
        \item $\mathrm{UCB}_i \leq \mathrm{UCB}_*$ (for maximization)
        \item $\mathrm{LCB}_i \geq \mathrm{LCB}_*$ (for minimization)
       \end{itemize}
       This results in $2\delta \leq \Delta_i$.
 \item Replace the $\delta$ with the exploration term.
       For example, in UCB1, $\delta=\sqrt{\frac{2\log T}{n_i}}$.
 \item Derive the lower bound $L$ for $n_i$ from $2\delta \leq \Delta_i$.
 \item Find the upper-bound of the probability of selecting a sub-optimal arm $i$.
       This is typically done by a union-bound argument.
 \item Derive the upper bound of the expected number of pulls $\E[n_i]$ of a suboptimal arm $i$
       using a triple loop summation.
       This is typically the heaviest part that needs mathematical tricks.
       The tricks do not seem generally transferable between approaches.
 \item Finally,
       derive an upper bound of the regret $TX_* - \sum_{i=1}^K X_i \E[n_i]$ by
       \begin{align*}
        TX_* - \sum_{i=1}^K X_i \E[n_i]
        &=\sum_{i=1}^K (X_* - X_i) \E[n_i]
        =\sum_{i=1}^K \Delta_i \E[n_i].
       \end{align*}
\end{enumerate}

\subsection{The Proof of UCB1 (Tutorial Example)}
\label{app:ucb1-proof}

We prove the logarithmic upper bound of the cumulative regret of the UCB1 $\hat{\mu}_i-c\sqrt{\frac{2\log T}{n_i}}$
where $\hat{\mu}_i$ is the empirical mean of samples from arm $i$,
$n_i$ is the number of pulls from arm $i$,
and $T=\sum_{i=1}^K n_i$ is the total pulls from all $K$ arms.

\begin{enumerate}
 \item UCB1 assumes a reward distribution with a known bound.
       For such a distribution, we can use Hoeffding's inequality.
       Given RVs $\rx_1\ldots \rx_n$, where $\rx_i\in [l_i, u_i]$, and their sum $\rS_n=\sum_{i=1}^n \rx_i$,
       \begin{align*}
        P(|\rS_n-\E[\rS_n]|\geq \epsilon)&\textstyle\leq 2\exp - \frac{2\epsilon^2}{\sum_{i=1}^{n} (u_i-l_i)^2}.
       \end{align*}
       Using $\delta=\frac{\epsilon}{n}$ and $\mu_n=\frac{\rS_n}{n}$,
       \begin{align*}
        P(|\mu_n-\E[\mu_n]\geq \delta|)\leq 2\exp - \frac{2n^2\delta^2}{\sum_{i=1}^{n} (u_i-l_i)^2}.
       \end{align*}
       UCB1 assumes $\rx_i$ are \iid copies, thus $\forall i; u_i-l_i=c$.
       \begin{align*}
        P(|\mu_n-\E[\mu_n]\geq \delta|)\leq 2\exp - \frac{2n^2\delta^2}{nc^2} = 2\exp - \frac{2n\delta^2}{c^2}.
       \end{align*}
 \item Expanding the two-sided error:
       \[
       \delta \geq \mu_n-\E[\mu_n] \geq -\delta.
       \]
 \item Changing the notation:
       \[
       \delta \geq \hat{\mu}_i-\mu_i \geq -\delta.
       \]
 \item Adding $\mu_i-\delta$ to both sides,
       \[
        \mu_i \geq \hat{\mu}_i-\delta = \mathrm{LCB}_{i}(T,n_i) \geq \mu_i-2\delta.
       \]
       Substituting $i=*$ (optimal arm), the first inequality is
       \[
        \mu_* \geq \hat{\mu}_*-\delta = \mathrm{LCB}_{*}(T,n_*).
       \]
       \textbf{Assuming} $2\delta\leq \Delta_i = \mu_i-\mu_*$,
       the second inequality is
       \[
       \mathrm{LCB}_{i}(T,n_i) \geq \mu_i-2\delta \geq \mu_i-\Delta_i = \mu_*.
       \]
       Therefore
       \[
       \mathrm{LCB}_{i}(T,n_i) \geq \mu_* \geq \mathrm{LCB}_{*}(T,n_*).
       \]
 \item Let $\delta=c\sqrt{\frac{2\log T}{n_i}}$.
       Then
       \begin{align*}
        P(\mu_{n_i}-\E[\mu_{n_i}]\geq \delta)\leq \exp - \frac{2n_i c^2\frac{2\log T}{{n_i}}}{c^2} = T^{-4}.
       \end{align*}
 \item From $2\delta\leq \Delta_i$, considering $n_i$ is an integer,
       \begin{align*}
        2c\sqrt{\frac{2\log T}{n_i}} \leq \Delta_i
        &\iff 4c^2 \frac{2\log T}{n_i} \leq \Delta_i^2\\
        &\hspace{-2em}\iff \frac{8c^2\log T}{\Delta_i^2} \leq \left\lceil\frac{8c^2\log T}{\Delta_i^2}\right\rceil=L\leq n_i.
       \end{align*}
 \item $\mathrm{LCB}_{i}(T,n_i) \geq \mu_* \geq \mathrm{LCB}_{*}(T,n_*)$
       does not hold when either inequality does not hold.
       $\mathrm{LCB}_{i}(T,n_i) \geq \mu_*$ does not hold with probability less than $T^{-4}$.
       $\mu_* \geq \mathrm{LCB}_{i}(T,n_*)$ does not hold with probability less than $T^{-4}$.
       Thus, by union-bound (probability of disjunctions),
       \[
       P(\mathrm{LCB}_{i}(T,n_i) \leq \mathrm{LCB}_{*}(T,n_*))\leq 2T^{-4}.
       \]

 \item Assume we followed the UCB1 strategy,
       i.e., we pulled the arm that minimizes the LCB.
       The expected number of pulls $\E[n_i]$ from a suboptimal arm $i$ is as follows.
       Note that for $K$ arms, every arm is at least pulled once.
       \begin{align*}
        \E[n_i]&=1+\sum_{t=K+1}^{T} P\parens{\text{$i$ is pulled at time $t$}}\\
        &\leq L +\sum_{t=K+1}^{T} P\parens{\text{$i$ is pulled at time $t$}\land n_i>L}\\
        &= L +\sum_{t=K+1}^{T} P\parens{\forall j; \mathrm{LCB}_j(t,n_j)\geq \mathrm{LCB}_i(t,n_i)}\\
        &\leq L +\sum_{t=K+1}^{T} P\parens{\mathrm{LCB}_*(t,n_*)\geq \mathrm{LCB}_i(t,n_i)}\\
        &\leq L +\sum_{t=K+1}^{T} P\parens{\exists u,v; \mathrm{LCB}_*(t,u)\geq \mathrm{LCB}_i(t,v)}\\
        &\leq L + \sum_{t=K+1}^{T} \sum_{u=1}^{t-1} \sum_{v=L}^{t-1}  P(\mathrm{LCB}_*(t,u)\geq \mathrm{LCB}_i(t,v))\\
        &\leq L + \sum_{t=K+1}^{T} \sum_{u=1}^{t-1} \sum_{v=L}^{t-1} 2t^{-4}\\
        &\leq L + \sum_{t=1}^{\infty} \sum_{u=1}^{t} \sum_{v=1}^{t} 2t^{-4}
        = L + \sum_{t=1}^{\infty} t^2 \cdot 2t^{-4}\\
        &= L + 2 \sum_{t=1}^{\infty} t^{-2}
        = L + 2\cdot \frac{\pi}{6} = L+\frac{\pi}{3}\\
        &\leq c^2 \frac{8\log T}{\Delta_i^2}+1 + \frac{\pi}{3} \quad \because \lceil x \rceil\leq x+1
       \end{align*}
 \item The regret is
       \begin{align*}
        T\mu_* - \sum_{i=1}^K \mu_i \E[n_i]
        &=\sum_{i=1}^K (\mu_* - \mu_i) \E[n_i]
        =\sum_{i=1}^K \Delta_i \E[n_i]\\
        &\leq \sum_{i=1}^K \Delta_i \parens{c^2 \frac{8\log T}{\Delta_i^2}+1 + \frac{\pi}{3}}\\
        &\leq \sum_{i=1}^K \parens{c^2 \frac{8\log T}{\Delta_i}+ \parens{1 + \frac{\pi}{3}}\Delta_i}.
       \end{align*}
\end{enumerate}

\section{The Proof of UCB1-Normal2}
\label{app:ucb-normal2-proof}

Our analysis begins with a definition of Sub-Gaussian distributions.

\begin{defi}
 \citep[Proposition 2.5.2, (iv)]{vershynin2018high}
 A distribution $p(x)$ is \emph{sub-Gaussian} when
 \[
 \exists t>0; \E[\exp x^2/t^2]<2.
 \]
\end{defi}

\begin{theo}
 A Gaussian distribution with 0-mean $\N(0,\sigma^2)$ (without loss of generality) is sub-Gaussian.
\end{theo}

\begin{proof}
 \[
  p(x)=\N(0,\sigma^2)=\frac{1}{\sqrt{2\pi\sigma^2}} \exp - \frac{x^2}{2\sigma^2}.
 \]
 \begin{align*}
  &\E[\exp x^2/t^2]
  = \int_{\R} \exp \frac{x^2}{t^2} \frac{1}{\sqrt{2\pi\sigma^2}} \exp - \frac{x^2}{2\sigma^2} dx\\
  &= \frac{1}{\sqrt{2\pi\sigma^2}} \int_{\R} \exp - x^2\parens{\frac{1}{2\sigma^2}-\frac{1}{t^2}} dx\\
  &= \frac{1}{\sqrt{2\pi\sigma^2}} \int_{\R} \exp - \frac{x^2}{C^2} dx\\
  &= \frac{1}{\sqrt{2\pi\sigma^2}} \int_{\R} \exp - y^2 Cdy\quad \parens{\frac{x}{C}=y \iff dx=Cdy}\\
  &= \frac{C}{\sqrt{2\pi\sigma^2}} \sqrt{\pi}\\
  &= \frac{C}{\sqrt{2\sigma^2}}.
 \end{align*}
 Where
 \begin{align*}
  \frac{1}{C^2} &= \frac{1}{2\sigma^2}-\frac{1}{t^2}\\
  \iff
  C^2 &= \frac{2\sigma^2t^2}{t^2-2\sigma^2}.
 \end{align*}
 To show $\E[\exp x^2/t^2]<2$,
 \begin{align*}
  \E[\exp x^2/t^2] &=\frac{C}{\sqrt{2\sigma^2}} = \sqrt{\frac{t^2}{t^2-2\sigma^2}}<2,\\
  &\iff
  t^2<4 (t^2-2\sigma^2),\\
  &\iff
  \frac{8}{3}\sigma^2 < t^2.
 \end{align*}
\end{proof}

\begin{defi}
 For a sub-Gaussian RV $x$, the \emph{sub-Gaussian norm} is defined as
 \[
 ||x||=\inf \braces{t>0\mid \E[\exp x^2/t^2]<2}.
 \]
\end{defi}
\begin{corollary}
 For $p(x)=\N(0,\sigma^2)$, $||x||=\sqrt{\frac{8}{3}}\sigma$.
\end{corollary}

Next, we review the general Hoeffding's inequality for sub-Gaussian distributions \citep[Theorem 2.6.2]{vershynin2018high}.

\begin{theo}
 For independent sub-Gaussian RVs $x_1,\ldots,x_n$,
 let their sum be $S_n=\sum_{i=1}^n x_i$.
 Then, for any $\epsilon>0$,
 \begin{align*}
  \Pr(|S_n - \E[S_n]| \leq \epsilon) &\geq 2\exp - \frac{\epsilon^2}{\sum_{i=1}^n ||x_i||^2},\\
  \Pr(S_n - \E[S_n] \leq \epsilon) &\geq \exp - \frac{\epsilon^2}{\sum_{i=1}^n ||x_i||^2},\\
  \Pr(\E[S_n] - S_n \leq \epsilon) &\geq \exp - \frac{\epsilon^2}{\sum_{i=1}^n ||x_i||^2}.
 \end{align*}
 (Two-sided bounds and one-sided upper/lower bounds, respectively.)
\end{theo}

We then prove UCB1-Normal2's regret, following the tutorial steps in the previous section.

\begin{enumerate}
 \item According to Hoeffding's inequality for sub-Gaussian
       RVs $X_1\ldots X_n$ and their sum $S_n=\sum_{i=1}^n X_i$,
       \begin{align*}
        P(|S_n-\E[S_n]\geq \epsilon|)&\textstyle\leq \exp - \frac{\epsilon^2}{\sum_{i=1}^{n} ||X_i||^2}.
       \end{align*}
       Using $\delta=\frac{\epsilon}{n}$,
       \begin{align*}
        P(|\mu_n-\E[\mu_n]\geq \delta|)\leq\textstyle \exp - \frac{n^2\delta^2}{\sum_{i=1}^{n} ||X_i||^2}.
       \end{align*}
       We assume $X_i=\N(\mu,\sigma^2)$, thus $||X_i||^2=\frac{8}{3}\sigma^2$.
       \begin{align*}
        P(|\mu_n-\E[\mu_n]\geq \delta|)\leq\textstyle \exp - \frac{3n^2\delta^2}{8n\sigma^2} = \exp - \frac{3n\delta^2}{8\sigma^2}.
       \end{align*}
 \item Same as UCB1.
 \item Same as UCB1.
 \item Same as UCB1.
 \item Let $\delta=\hat{\sigma}\sqrt{\log T}$.
       Then
       \begin{align*}
        P(A:\mu_{n_i}-\E[\mu_{n_i}]\geq \delta) &\leq \exp - \frac{3{n_i} \hat{\sigma}^2\log T}{8\sigma^2} \\
        &= T^{-\frac{3{n_i}\hat{\sigma}^2}{8\sigma^2}}.
       \end{align*}
       The trick starts here.
       The formula above is problematic because we do not know the true variance $\sigma^2$.
       However, if event $B:\frac{{n_i}\hat{\sigma}^2}{\sigma^2}\geq X$ holds for some $X>0$,
       we have
       \begin{align*}
        \textstyle
        T^{-\frac{3{n_i}\hat{\sigma}^2}{8\sigma^2}}\leq T^{-\frac{3}{8}X}.
       \end{align*}
       One issue with this approach is that the two events $A,B$ may be correlated.
       To address the issue, we further upper-bound the probability by union-bound.
       Let $P(B)=\alpha$ which is close to 1.
       Then
       \begin{align*}
        P(\lnot (A\land B))
        =P(\lnot A \lor \lnot B)
        & \leq P(\lnot A) + P(\lnot B).\\
        1-P(A\land B) &\leq 1-P(A) + P(\lnot B).\\
        P(A)&\leq P(A\land B)+P(\lnot B).\\
        \therefore
        P(\mu_{n_i}-\E[\mu_{n_i}]\geq \delta)
        &\leq T^{-\frac{3}{8}X}+1-\alpha.
       \end{align*}
       We next obtain $X$ that satisfies $P(B)=\alpha$.
       We use the fact that $\frac{{n_i}\hat{\sigma}^2}{\sigma^2}$ follows
       a Chi-Squared distribution $\chi^2({n_i})$ with a degree of freedom ${n_i}$.
       Then $X=\chi^2_{1-\alpha,n_i}$,
       the upper-tail critical value of $\chi^2$ distribution with degree of freedom $n_i$ and significance level $\alpha$,
       because
       \begin{align*}
        P(\lnot B)
        &=P(\frac{{n_i}\hat{\sigma}^2}{\sigma^2}<\chi^2_{1-\alpha,n_i})
        \\
        &=\chi^2(\frac{{n_i}\hat{\sigma}^2}{\sigma^2}<\chi^2_{1-\alpha,n_i}\mid {n_i})
        =1-\alpha
        .
       \end{align*}

 \item From $2\delta\leq \Delta_i$, assuming $n_i$ is an integer and $n_i\geq 2$,
       \begin{align*}
        \Delta_i^2
        &\geq 2\hat{\sigma}^2 \log T = \frac{2n_i\hat{\sigma}^2 \log T}{n_i}
         \geq \frac{2 \sigma^2 \chi^2_{1-\alpha,n_i} \log T}{n_i}\\
        &\geq \frac{2 \sigma^2 \chi^2_{1-\alpha,2} \log T}{n_i}
         =    \frac{-4 \sigma^2 \log \alpha \log T}{n_i}.\\
        \therefore
        n_i &\geq \left\lceil\frac{-4 \sigma^2 \log \alpha \log T}{\Delta^2_i}\right\rceil=L\geq \frac{-4 \sigma^2 \log \alpha \log T}{\Delta^2_i}.
       \end{align*}
       Note that we used the fact that $\chi^2_{1-\alpha,n}$ is monotonically increasing for $n$,
       therefore $\chi^2_{1-\alpha,n}\geq \chi^2_{1-\alpha,2}$ ($n_i\geq 2$),
       and that
       $\chi^2_{1-\alpha,2}=-2\log \alpha$:
       \begin{align*}
        1-\alpha
        &=\chi^2(X<\chi^2_{1-\alpha,n}\mid n=2)\\
        &=\frac{\gamma(\frac{2}{2},\frac{\chi^2_{1-\alpha,2}}{2})}{\Gamma(\frac{2}{2})}=1-e^{-\frac{\chi^2_{1-\alpha,2}}{2}}.
       \end{align*}
       where $\gamma$ and $\Gamma$ are (incomplete) Gamma functions.
 \item Using the same union-bound argument used in UCB1,
  \begin{align*}
   P(\mathrm{LCB}_{i}(T,n_i) &\leq \mathrm{LCB}_{*}(T,n_*))\\
   &\leq 2(T^{-\chi^2_{1-\alpha,n_i}}+1-\alpha).
  \end{align*}

 \item Assume we followed the UCB1-Normal2 strategy.
       We use the same argument as UCB1.
       Assume we pull each arm at least $M$ times in the beginning and $M\leq L$.
       \begin{align*}
        \E[n_i]
        &\leq L +\sum_{t=K+1}^{T} P\parens{\exists u,v; \mathrm{LCB}_*(t,u)\geq \mathrm{LCB}_i(t,v)}\\
        &\leq L +\sum_{t=K+1}^{T} \sum_{u=1}^{t-1} \sum_{v=L}^{t-1} 2(t^{-\frac{3}{8}\chi^2_{1-\alpha,v}}+1-\alpha)\\
        &\leq L +\sum_{t=K+1}^{T} \sum_{u=1}^{t-1} \sum_{v=L}^{t-1} 2(t^{-\frac{3}{8}\chi^2_{1-\alpha,M}}+1-\alpha) \\
        &\leq L +\sum_{t=K+1}^{T} \sum_{u=1}^{t} \sum_{v=1}^{t} 2(t^{-\frac{3}{8}\chi^2_{1-\alpha,M}}+1-\alpha) \\
        &=    L +\sum_{t=K+1}^{T} 2(t^{2-\frac{3}{8}\chi^2_{1-\alpha,M}}+(1-\alpha) t^2)\\
        &\leq L +2\sum_{t=1}^{\infty} t^{2-\frac{3}{8}\chi^2_{1-\alpha,M}} + 2(1-\alpha) \sum_{t=1}^{T} t^2\\
        &=    L + 2C + 2(1-\alpha) \frac{T(T+1)(2T+1)}{6} \\
        &\leq \frac{-4\sigma^2\log \alpha \log T}{\Delta^2_i} + 1 \quad (\because \lceil x \rceil\leq x+1)\\
        &\quad + 2C + \frac{(1-\alpha) T(T+1)(2T+1)}{3}.
       \end{align*}
       $C$ is a convergent series when
       \begin{align*}
        2-\frac{3}{8}\chi^2_{1-\alpha,M}<-1 \ \iff \ 8<\chi^2_{1-\alpha,M}.
       \end{align*}
       You can look up the value of $M$ that guarantees this condition
       from a numerically computed, so-called \emph{$\chi^2$-table} (\reftbl{tab:chi-square-table}).
       For example, with $\alpha=0.99$, $8<\chi^2_{0.01,M}$, thus $M\geq 2$, and
       with $\alpha=0.9$, $8<\chi^2_{0.1,M}$, thus $M\geq 5$.
       However, the value of $\alpha$ depends on the problem and is unknown prior to solving the problem.
 \item Omitted.
\end{enumerate}

\begin{table*}
 \centering
 \begin{tabular}{|*{11}{c|}}
\toprule
\backslashbox{$n$}{$\alpha$}
   &0.995 &0.99 &0.975 &0.95 &0.90 &0.10 &0.05 &0.025 &0.01 &0.005 \\\midrule
1 &--- &--- &0.001 &0.004 &0.016 &2.706 &3.841 &5.024 &6.635 &7.879 \\
2 &0.010 &0.020 &0.051 &0.103 &0.211 &4.605 &5.991 &7.378 &9.210 &10.597 \\
3 &0.072 &0.115 &0.216 &0.352 &0.584 &6.251 &7.815 &9.348 &11.345 &12.838 \\
4 &0.207 &0.297 &0.484 &0.711 &1.064 &7.779 &9.488 &11.143 &13.277 &14.860 \\
5 &0.412 &0.554 &0.831 &1.145 &1.610 &9.236 &11.070 &12.833 &15.086 &16.750 \\
6 &0.676 &0.872 &1.237 &1.635 &2.204 &10.645 &12.592 &14.449 &16.812 &18.548 \\
7 &0.989 &1.239 &1.690 &2.167 &2.833 &12.017 &14.067 &16.013 &18.475 &20.278 \\
8 &1.344 &1.646 &2.180 &2.733 &3.490 &13.362 &15.507 &17.535 &20.090 &21.955 \\
9 &1.735 &2.088 &2.700 &3.325 &4.168 &14.684 &16.919 &19.023 &21.666 &23.589 \\
10 &2.156 &2.558 &3.247 &3.940 &4.865 &15.987 &18.307 &20.483 &23.209 &25.188 \\
11 &2.603 &3.053 &3.816 &4.575 &5.578 &17.275 &19.675 &21.920 &24.725 &26.757 \\
12 &3.074 &3.571 &4.404 &5.226 &6.304 &18.549 &21.026 &23.337 &26.217 &28.300 \\
13 &3.565 &4.107 &5.009 &5.892 &7.042 &19.812 &22.362 &24.736 &27.688 &29.819 \\
14 &4.075 &4.660 &5.629 &6.571 &7.790 &21.064 &23.685 &26.119 &29.141 &31.319 \\
15 &4.601 &5.229 &6.262 &7.261 &8.547 &22.307 &24.996 &27.488 &30.578 &32.801 \\
16 &5.142 &5.812 &6.908 &7.962 &9.312 &23.542 &26.296 &28.845 &32.000 &34.267 \\
17 &5.697 &6.408 &7.564 &8.672 &10.085 &24.769 &27.587 &30.191 &33.409 &35.718 \\
18 &6.265 &7.015 &8.231 &9.390 &10.865 &25.989 &28.869 &31.526 &34.805 &37.156 \\
19 &6.844 &7.633 &8.907 &10.117 &11.651 &27.204 &30.144 &32.852 &36.191 &38.582 \\
20 &7.434 &8.260 &9.591 &10.851 &12.443 &28.412 &31.410 &34.170 &37.566 &39.997 \\
21 &8.034 &8.897 &10.283 &11.591 &13.240 &29.615 &32.671 &35.479 &38.932 &41.401 \\
22 &8.643 &9.542 &10.982 &12.338 &14.041 &30.813 &33.924 &36.781 &40.289 &42.796 \\
23 &9.260 &10.196 &11.689 &13.091 &14.848 &32.007 &35.172 &38.076 &41.638 &44.181 \\
24 &9.886 &10.856 &12.401 &13.848 &15.659 &33.196 &36.415 &39.364 &42.980 &45.559 \\
25 &10.520 &11.524 &13.120 &14.611 &16.473 &34.382 &37.652 &40.646 &44.314 &46.928 \\
26 &11.160 &12.198 &13.844 &15.379 &17.292 &35.563 &38.885 &41.923 &45.642 &48.290 \\
27 &11.808 &12.879 &14.573 &16.151 &18.114 &36.741 &40.113 &43.195 &46.963 &49.645 \\
28 &12.461 &13.565 &15.308 &16.928 &18.939 &37.916 &41.337 &44.461 &48.278 &50.993 \\
29 &13.121 &14.256 &16.047 &17.708 &19.768 &39.087 &42.557 &45.722 &49.588 &52.336 \\
30 &13.787 &14.953 &16.791 &18.493 &20.599 &40.256 &43.773 &46.979 &50.892 &53.672 \\
40 &20.707 &22.164 &24.433 &26.509 &29.051 &51.805 &55.758 &59.342 &63.691 &66.766 \\
50 &27.991 &29.707 &32.357 &34.764 &37.689 &63.167 &67.505 &71.420 &76.154 &79.490 \\
60 &35.534 &37.485 &40.482 &43.188 &46.459 &74.397 &79.082 &83.298 &88.379 &91.952 \\
70 &43.275 &45.442 &48.758 &51.739 &55.329 &85.527 &90.531 &95.023 &100.425 &104.215 \\
80 &51.172 &53.540 &57.153 &60.391 &64.278 &96.578 &101.879 &106.629 &112.329 &116.321 \\
90 &59.196 &61.754 &65.647 &69.126 &73.291 &107.565 &113.145 &118.136 &124.116 &128.299 \\
100 &67.328 &70.065 &74.222 &77.929 &82.358 &118.498 &124.342 &129.561 &135.807 &140.169   \\\bottomrule
 \end{tabular}
 \caption{$\chi^2_{\alpha,n}$ table.}
 \label{tab:chi-square-table}
\end{table*}

\section{Statistics after Merging Datasets}
\label{app:backprop}

Backpropagation in MCTS requires
computing the statistics of the samples in the leaf nodes in a subtree of a parent node.
To avoid iterating over all leaves of each parent,
Backpropagation typically propagates the statistics from the immediate children.
This can be seen as merging multiple datasets
and compute the statistics of the merged dataset
from the statistics of multiple datasets.

In variance-based MCTS algorithms, both the mean and variance are backpropagated.
Given two sets of samples $X_1, X_2$,
each with an empirical mean $\mu_i$ and $n_i$ elements $(i\in\braces{1,2})$,
the empirical mean $\mu_{12}$ of $X_1\cup X_2$ is given by
\begin{align*}
 \mu_{12}
 &\textstyle
 = \frac{\sum_{x\in X_1} x + \sum_{x\in X_2} x}{n_1+n_2}
 = \frac{n_1\mu_1+n_2\mu_2}{n_1+n_2}
\end{align*}
We obtain NECs by iterating this process over a node's children,
although there is a more efficient, incremental method
for backpropagating a change in a single child (see appendix).
For the variance, we similarly merge the samples.
Given individual variances $\sigma^2_1$ and $\sigma^2_2$,
the variance $\sigma^2_{12}$ of $X_1\cup X_2$ (proof available in appendix) is:
\begin{align*}
 \sigma^2_{12}
 &\textstyle=\frac{n_1\sigma_1^2 + n_2\sigma_2^2 + \frac{n_1n_2}{n_1+n_2}(\mu_2-\mu_1)^2}{n_1+n_2}.
\end{align*}

Below, we show the formulae and the proofs for this method.

\begin{theo}[The empirical mean of merged datasets]
 Given two sets of samples $X_1, X_2$,
each with an empirical mean $\mu_i$ and $n_i$ elements $(i\in\braces{1,2})$,
the empirical mean $\mu_{12}$ of $X_1\cup X_2$ is given by
\begin{align*}
 \mu_{12} &= \frac{n_1\mu_1+n_2\mu_2}{n_1+n_2}.
 \\
 \text{Also,}\quad
 \mu_{12}&= \mu_1 + \frac{n_2}{n_1+n_2} (\mu_2-\mu_1).
\end{align*}
\end{theo}

\begin{proof}
 \begin{align*}
  \mu_{12}
  &= \frac{\sum_{x\in X_1} x + \sum_{x\in X_2} x}{n_1+n_2}
  = \frac{n_1\mu_1 + n_2\mu_2}{n_1+n_2}.
 \end{align*}
\end{proof}

\begin{theo}[The empirical variance of merged datasets]
 Given two sets of samples $X_1, X_2$,
 each with an empirical mean $\mu_i$, variance $\sigma^2_i$, and $n_i$ elements $(i\in\braces{1,2})$,
 and $n_{ij}=n_i+n_j$,
 the empirical variance $\mu_{12}$ of $X_1\cup X_2$ is given by
\begin{align*}
 \sigma^2_{12}
 &=\frac{n_1\sigma_1^2 + n_2\sigma_2^2 + \frac{n_1n_2}{n_1+n_2}(\mu_2-\mu_1)^2}{n_1+n_2}.
\end{align*}
\end{theo}

\begin{proof}
 \begin{align*}
  \sigma^2_{12}
  &= \frac{\sum_{x\in X_1} (x-\mu_{12})^2 + \sum_{x\in X_2} (x-\mu_{12})^2}{n_1+n_2}.
 \end{align*}
 \begin{align*}
  \sum_{x\in X_1}& (x-\mu_{12})^2\\
  =&\sum_{x\in X_1} (x- (\mu_1 + \frac{n_2}{n_1+n_2} (\mu_2-\mu_1)))^2\\
  =&\sum_{x\in X_1} ((x- \mu_1) - \frac{n_2}{n_1+n_2} (\mu_2-\mu_1)))^2\\
  =&\sum_{x\in X_1} (x- \mu_1)^2
  -2 \sum_{x\in X_1} (x- \mu_1)\frac{n_2}{n_1+n_2} (\mu_2-\mu_1) \\
  &+ \sum_{x\in X_1}\parens{\frac{n_2}{n_1+n_2} (\mu_2-\mu_1)}^2\\
  =&\ n_1\sigma_1^2 -2\cdot 0 + n_1 (\frac{n_2}{n_1+n_2} (\mu_2-\mu_1))^2.\\
  \therefore
  (n_1&\ +n_2)\sigma^2_{12}\\
  =&\ n_1\sigma_1^2 + n_1 (\frac{n_2}{n_1+n_2} (\mu_2-\mu_1))^2\\
  &+ n_2\sigma_2^2 + n_2 (\frac{n_1}{n_1+n_2} (\mu_1-\mu_2))^2\\
  =&\ n_1\sigma_1^2 + n_2\sigma_2^2 + \frac{n_1n_2^2+n_2n_1^2}{(n_1+n_2)^2}(\mu_2-\mu_1)^2\\
  =&\ n_1\sigma_1^2 + n_2\sigma_2^2 + \frac{n_1n_2}{n_1+n_2}(\mu_2-\mu_1)^2.
 \end{align*}
\end{proof}

\section{Statistics after Retracting a Dataset}

In the backpropagation step of MCTS, typically,
only a few children of an intermediate node update their statistics (most often a single children).
To compute the updated statistics efficiently,
we could compute them by
retracting the old data of the child(ren) from the merged data and
merging the new data for the child(ren),
rather than iterating over the children to merge everything from scratch.
This can impact the performance when the number of children / the branching factor is high.

\begin{theo}[The empirical mean after retracting a dataset]
 Assume samples $X_1, X_2$
 with empirical means $\mu_{i}$ and number of elements $n_i$ ($i\in\braces{1,2}$).
 Let their union be $X_{12}=X_1\cup X_2$,
 its empirical means $\mu_{12}$, and its number of elements $n_{12}=n_1+n_2$.
 $\mu_{1}$ is given by
\begin{align*}
 \mu_{1} &= \frac{n_{12}\mu_{12} - n_2\mu_2}{n_{12}-n_2}.
\end{align*}
\end{theo}

\begin{theo}[The empirical variance after retracting a dataset]
 Assume samples $X_1, X_2$
 with
 empirical means $\mu_{i}$,
 empirical variance $\sigma^2_i$,
 and number of elements $n_i$ ($i\in\braces{1,2}$).
 Let their union be $X_{12}=X_1\cup X_2$,
 its empirical mean $\mu_{12}$,
 its empirical variance $\sigma^2_{12}$,
 and its number of elements $n_{12}=n_1+n_2$.
 $\sigma^2_{1}$ is given by
 $\mu_{12}$, $\mu_{2}$, $\mu_{1}$,
 $n_{1}$, $n_{2}$, $n_{12}$,
 $\sigma^2_{12}$, and $\sigma^2_{2}$ as follows.
\begin{align*}
 \sigma^2_{1}
 =\frac{1}{n_{1}}
 \parens{
 n_{12}\sigma^2_{12} - n_2\sigma_2^2 - \frac{n_2 n_{12}}{n_1} (\mu_{12} - \mu_2)^2
 }
\end{align*}
\end{theo}

\begin{proof}
 \begin{align*}
  n_{12}\sigma^2_{12}
  &=n_1\sigma_1^2 + n_2\sigma_2^2 + \frac{n_1n_2}{n_{12}}(\mu_2-\mu_1)^2.
  \\
  \therefore
  n_1\sigma^2_{1}
  &=
  n_{12}\sigma^2_{12} - n_2\sigma_2^2 - \frac{n_1n_2}{n_{12}}(\mu_2-\mu_1)^2.
  \\
  n_1\mu_1
  &=n_{12} \mu_{12} - n_2 \mu_2.
  \\
  \therefore
  \mu_1-\mu_2
  &=\frac{n_{12} \mu_{12} - n_2 \mu_2}{n_1}-\mu_2\\
  &=\frac{n_{12} \mu_{12} - n_2 \mu_2 - n_1 \mu_2}{n_1}\\
  &=\frac{n_{12} \mu_{12} - n_2 \mu_2 - (n_{12}-n_2) \mu_2}{n_1}\\
  &=\frac{n_{12}}{n_1} (\mu_{12} - \mu_2).
  \\
  \therefore
  n_1\sigma^2_{1}
  &=
  n_{12}\sigma^2_{12} - n_2\sigma_2^2 - \frac{n_1n_2}{n_{12}} \frac{n_{12}^2}{n_1^2} (\mu_{12} - \mu_2)^2\\
  &=
  n_{12}\sigma^2_{12} - n_2\sigma_2^2 - \frac{n_2 n_{12}}{n_1} (\mu_{12} - \mu_2)^2.
 \end{align*}
\end{proof}

\section{Further Results}

\subsection{Cumulative Histograms for All Heuristics and All Search Statistics}

\begin{figure*}[tb]
 \centering
  \begin{tabular}{cc|c}
   GUCT & GUCT* & \\\hline
   \includegraphics[valign=c,width=0.3\linewidth]{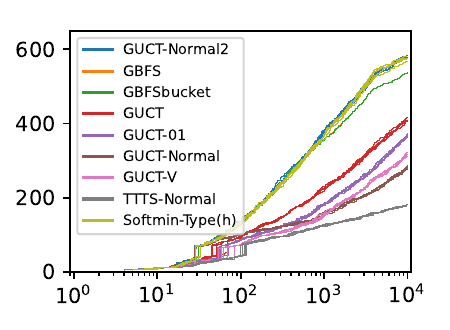}
   & \includegraphics[valign=c,width=0.3\linewidth]{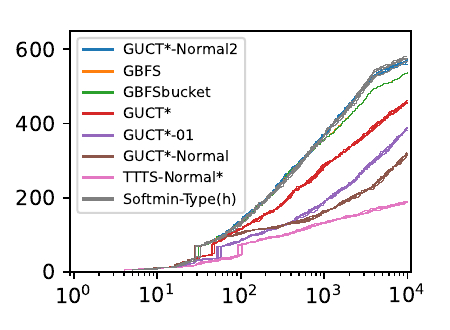}
       & \rotatebox[origin=c]{-90}{$\ff$}
           \\
   \includegraphics[valign=c,width=0.3\linewidth]{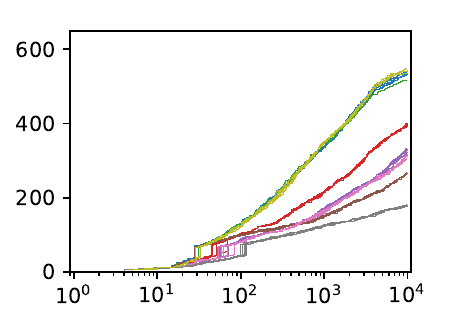}
   & \includegraphics[valign=c,width=0.3\linewidth]{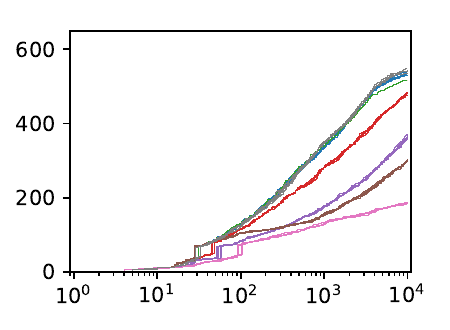}
       & \rotatebox[origin=c]{-90}{$\ad$}
           \\
   \includegraphics[valign=c,width=0.3\linewidth]{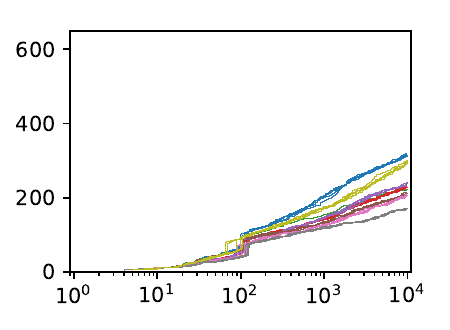}
   & \includegraphics[valign=c,width=0.3\linewidth]{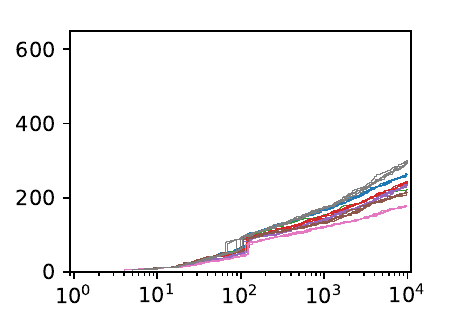}
       & \rotatebox[origin=c]{-90}{$\hmax$}
           \\
   \includegraphics[valign=c,width=0.3\linewidth]{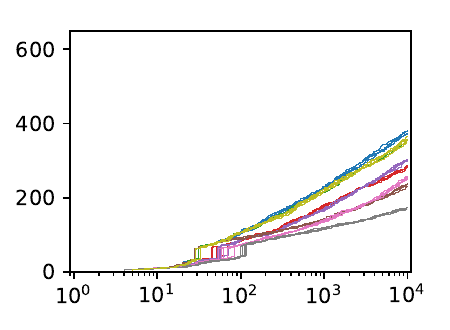}
   & \includegraphics[valign=c,width=0.3\linewidth]{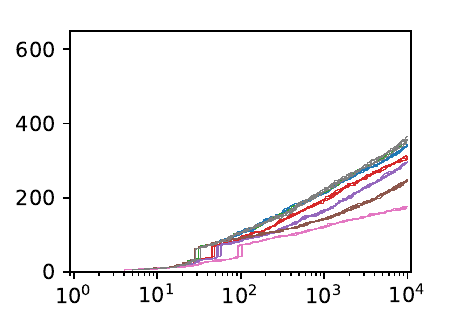}
       & \rotatebox[origin=c]{-90}{$\gc$}
  \end{tabular}
 \caption{
 The cumulative histogram of the number of problem instances solved ($y$-axis)
 below a certain number of node evaluations ($x$-axis, 10,000 nodes maximum).
 Each line represents a random seed.
 In algorithms with an exploration coefficient hyperparameter, we use $c=1.0$.
 The total numbers at the limit differ from those in other plots (this result does not limit the expansions or the runtime).
 }
 \label{fig:evaluation-histogram}
\end{figure*}

\begin{figure*}[tb]
 \centering
  \begin{tabular}{cc|c}
   GUCT & GUCT* & \\\hline
   \includegraphics[valign=c,width=0.3\linewidth]{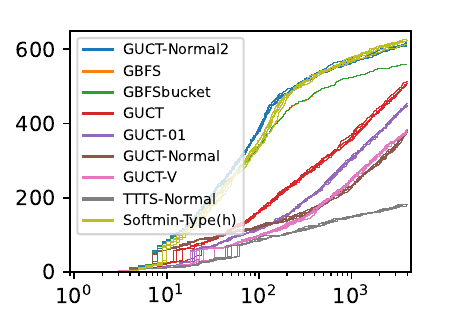}
   & \includegraphics[valign=c,width=0.3\linewidth]{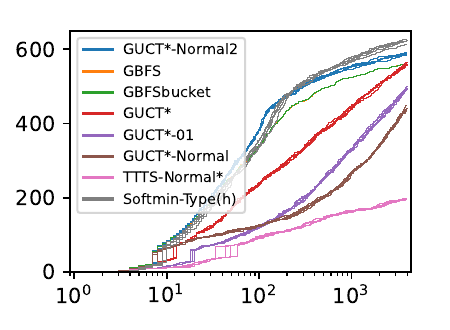}
       & \rotatebox[origin=c]{-90}{$\ff$}
           \\
   \includegraphics[valign=c,width=0.3\linewidth]{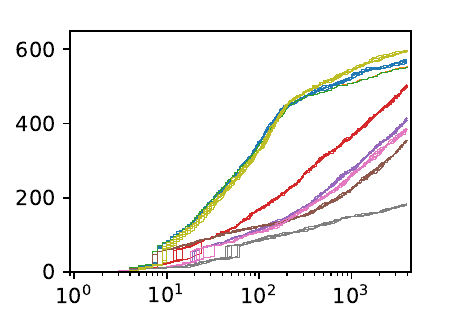}
   & \includegraphics[valign=c,width=0.3\linewidth]{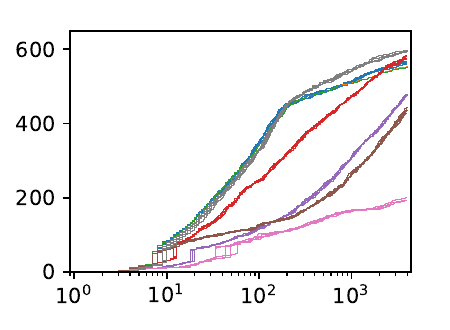}
       & \rotatebox[origin=c]{-90}{$\ad$}
           \\
   \includegraphics[valign=c,width=0.3\linewidth]{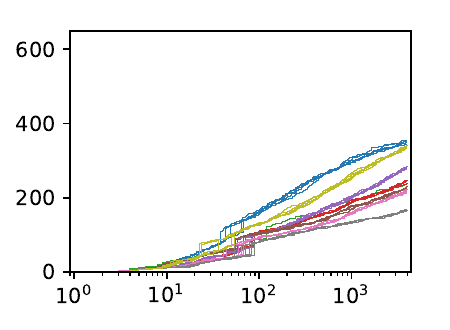}
   & \includegraphics[valign=c,width=0.3\linewidth]{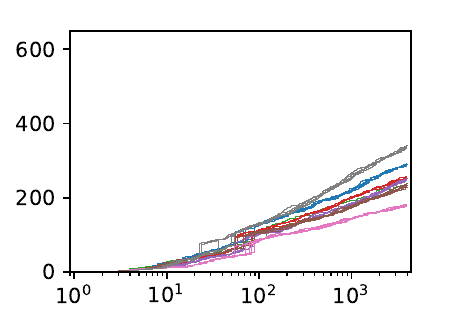}
       & \rotatebox[origin=c]{-90}{$\hmax$}
           \\
   \includegraphics[valign=c,width=0.3\linewidth]{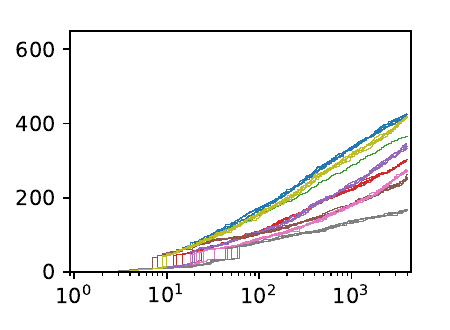}
   & \includegraphics[valign=c,width=0.3\linewidth]{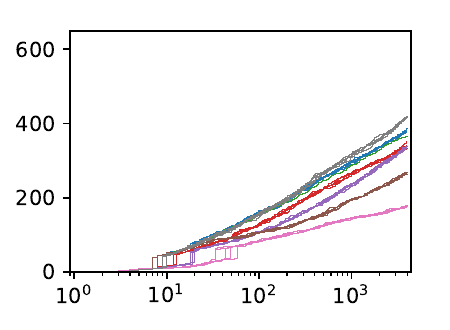}
       & \rotatebox[origin=c]{-90}{$\gc$}
  \end{tabular}
 \caption{
 The cumulative histogram of the number of problem instances solved ($y$-axis)
 below a certain number of node expansions ($x$-axis, 4,000 nodes maximum).
 Each line represents a random seed.
 In algorithms with an exploration coefficient hyperparameter, we use $c=1.0$.
 The total numbers at the limit differ from those in other plots (this result does not limit the evaluations or the runtime).
 }
 \label{fig:expansion-histogram}
\end{figure*}

\begin{figure*}[tb]
 \centering
  \begin{tabular}{cc|c}
   GUCT & GUCT* & \\\hline
   \includegraphics[valign=c,width=0.3\linewidth]{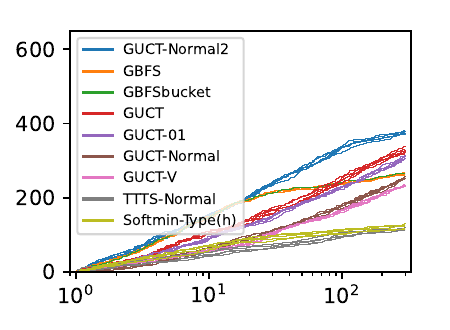}
   & \includegraphics[valign=c,width=0.3\linewidth]{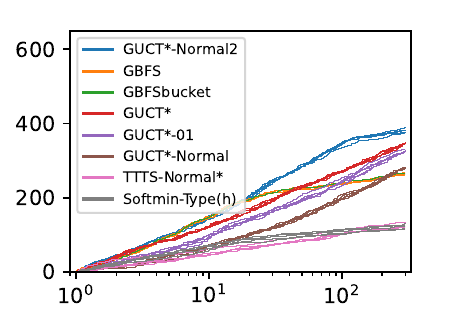}
       & \rotatebox[origin=c]{-90}{$\ff$}
           \\
   \includegraphics[valign=c,width=0.3\linewidth]{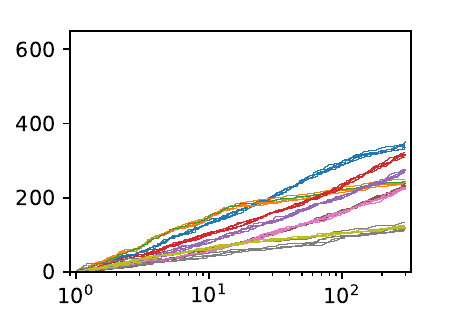}
   & \includegraphics[valign=c,width=0.3\linewidth]{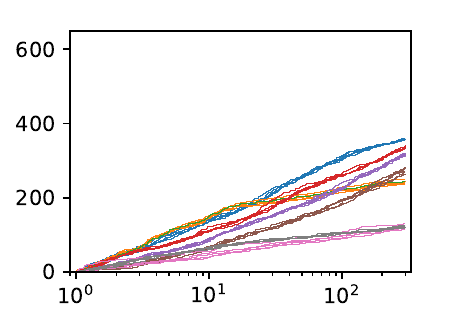}
       & \rotatebox[origin=c]{-90}{$\ad$}
           \\
   \includegraphics[valign=c,width=0.3\linewidth]{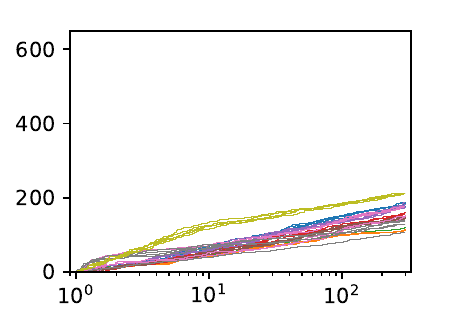}
   & \includegraphics[valign=c,width=0.3\linewidth]{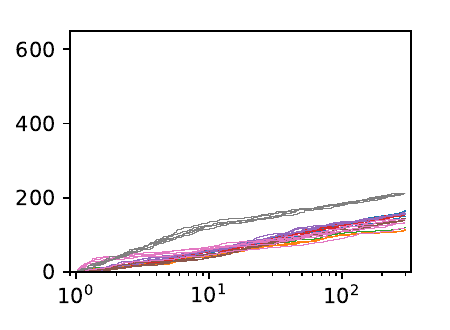}
       & \rotatebox[origin=c]{-90}{$\hmax$}
           \\
   \includegraphics[valign=c,width=0.3\linewidth]{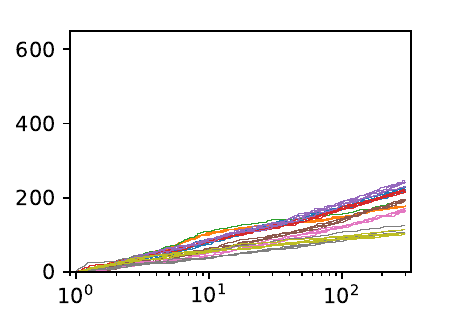}
   & \includegraphics[valign=c,width=0.3\linewidth]{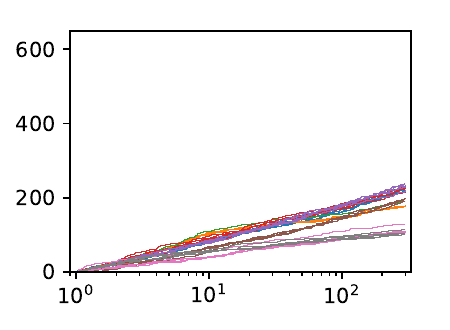}
       & \rotatebox[origin=c]{-90}{$\gc$}
  \end{tabular}
 \caption{
 The cumulative histogram of the number of problem instances solved ($y$-axis)
 below a certain runtime ($x$-axis, 300 seconds maximum).
 Each line represents a random seed.
 In algorithms with an exploration coefficient hyperparameter, we use $c=1.0$.
 The total numbers at the limit differ from those in other plots (this result does not limit the evaluations or the expansion).
 GBFS shows a unique slowdown, potentially due to the suboptimal heap-based open list implementation in Pyperplan,
 which is not necessarily a representative performance of GBFS in general (e.g., in Fast Downward, it can be implemented as a bucket-based open list).
 To reject this hypothesis, we also implemented a GBFS using bucket-based queue in Pyperplan,
 whose results are shown as GBFSbucket.
 The results indicate that GBFSbucket and GBFS has a similar runtime curve,
 indicating that the unique curve of GBFS is not due to the efficiency of open list insertion.
 }
 \label{fig:elapsed-histogram}
\end{figure*}

\refigs{fig:evaluation-histogram}{fig:elapsed-histogram} shows
the cumulative histogram of the number of instances solved by a particular evaluation/expansion/runtime.

\subsection{Deferred Heuristic Evaluation}

\begin{figure*}[p]
 \centering
  \begin{tabular}{ccc|c}
   evaluated & expanded & elapsed & $h$ \\\hline
    \includegraphics[valign=c,width=0.28\linewidth]{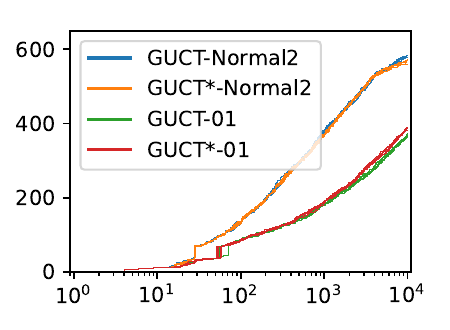}
   &\includegraphics[valign=c,width=0.28\linewidth]{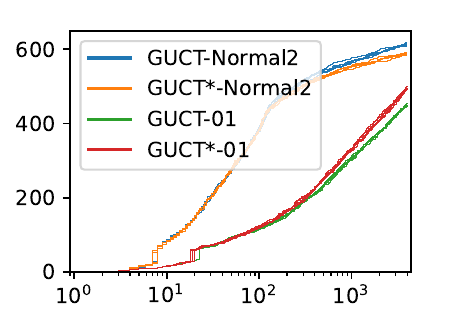}
   &\includegraphics[valign=c,width=0.28\linewidth]{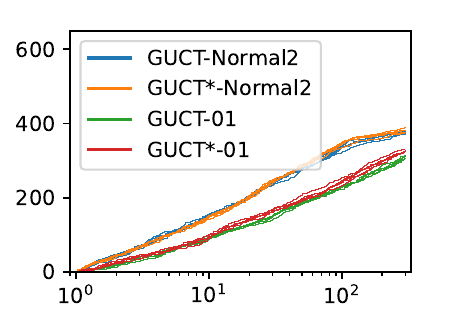}
   & \rotatebox[origin=c]{-90}{$\ff$}
       \\
    \includegraphics[valign=c,width=0.28\linewidth]{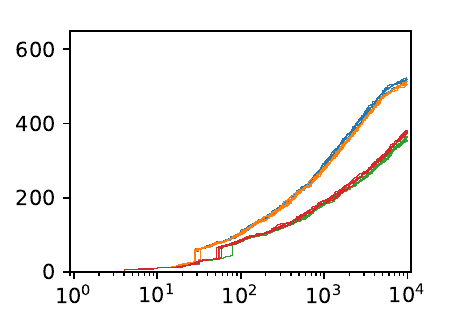}
   &\includegraphics[valign=c,width=0.28\linewidth]{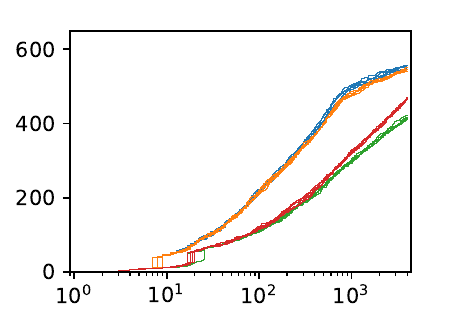}
   &\includegraphics[valign=c,width=0.28\linewidth]{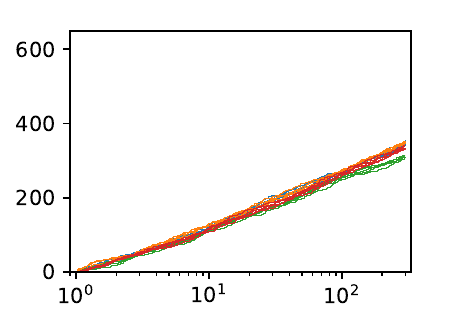}
   & \rotatebox[origin=c]{-90}{$\ff$+DE}
       \\
    \includegraphics[valign=c,width=0.28\linewidth]{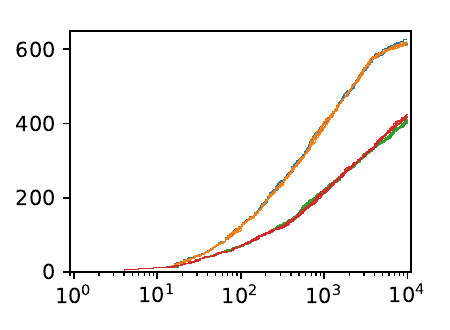}
   &\includegraphics[valign=c,width=0.28\linewidth]{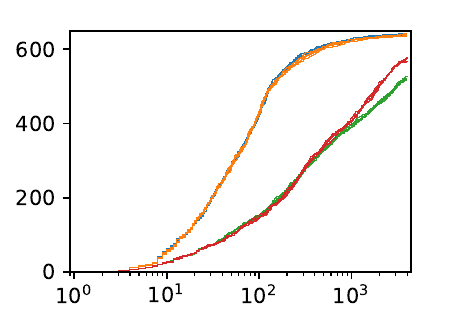}
   &\includegraphics[valign=c,width=0.28\linewidth]{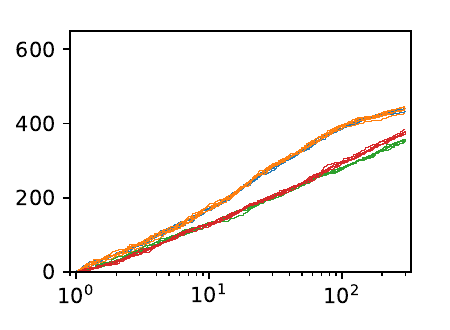}
   & \rotatebox[origin=c]{-90}{$\ff$+PO}
       \\
    \includegraphics[valign=c,width=0.28\linewidth]{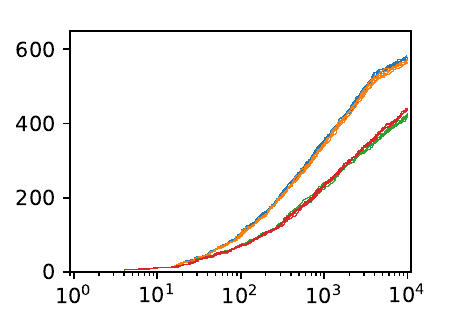}
   &\includegraphics[valign=c,width=0.28\linewidth]{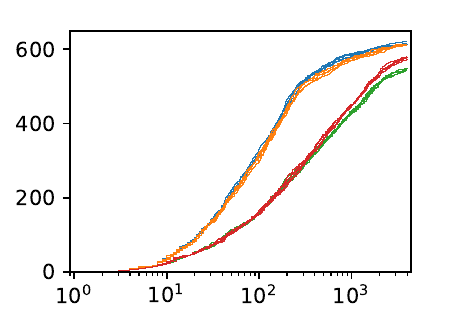}
   &\includegraphics[valign=c,width=0.28\linewidth]{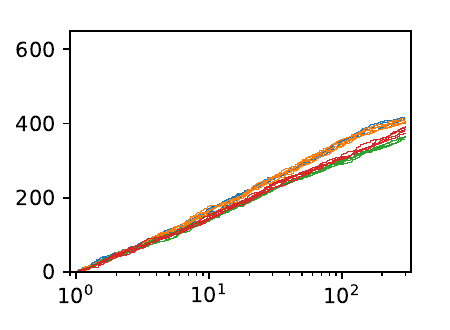}
   & \rotatebox[origin=c]{-90}{$\ff$+DE+PO}
  \end{tabular}
 \caption{
 The cumulative histogram of the number of problem instances solved ($y$-axis)
 below a certain evaluations/expansions/runtime ($x$-axis).
 Each line represents a random seed.
 In algorithms with an exploration coefficient hyperparameter, we use $c=1.0$.
 }
 \label{fig:histogram-depo}
\end{figure*}

\refig{fig:histogram-depo} shows
the cumulative histogram of the number of instances solved under a particular evaluation/expansion/runtime
by $\ff$ with/without DE, with/without PO.

\subsection{Comparing Different $c$ values}
\label{sec:different-c}

\begin{figure*}[tb]
 \centering
  \begin{tabular}{ccc|c}
   evaluated & expanded & elapsed & $h$ \\\midrule
   \includegraphics[valign=c,width=0.3\linewidth]{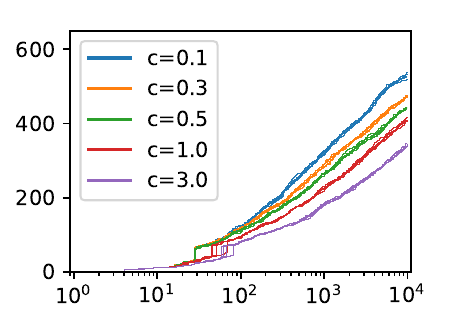}&
   \includegraphics[valign=c,width=0.3\linewidth]{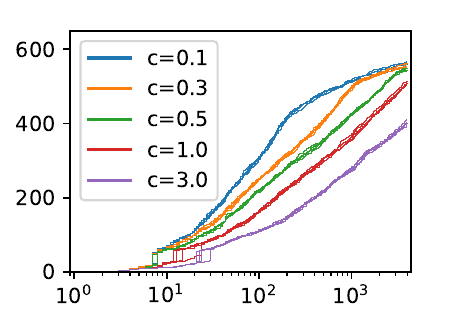}&
   \includegraphics[valign=c,width=0.3\linewidth]{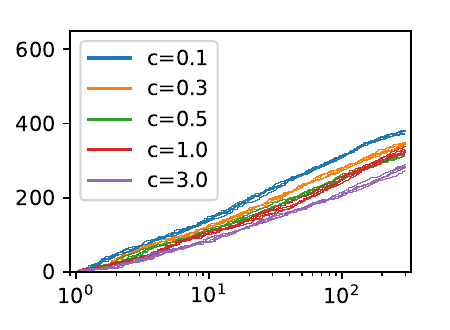}&
   \rotatebox[origin=c]{-90}{$\ff$}
       \\
   \includegraphics[valign=c,width=0.3\linewidth]{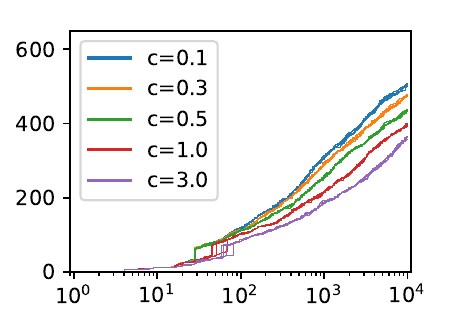}&
   \includegraphics[valign=c,width=0.3\linewidth]{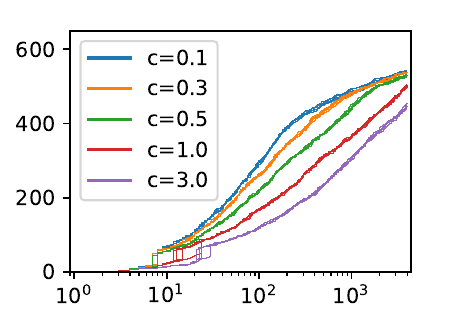}&
   \includegraphics[valign=c,width=0.3\linewidth]{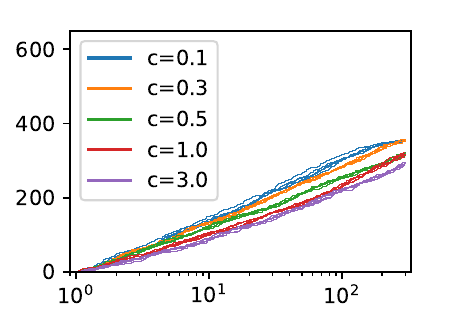}&
   \rotatebox[origin=c]{-90}{$\ad$}
       \\
   \includegraphics[valign=c,width=0.3\linewidth]{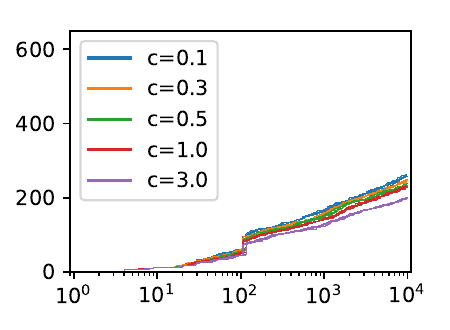}&
   \includegraphics[valign=c,width=0.3\linewidth]{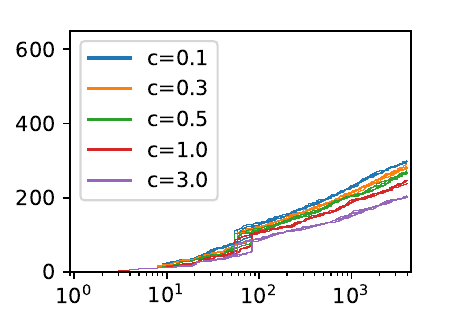}&
   \includegraphics[valign=c,width=0.3\linewidth]{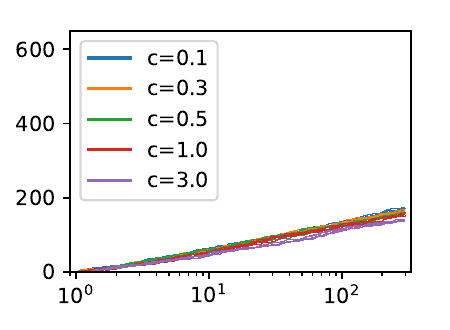}&
   \rotatebox[origin=c]{-90}{$\hmax$}
       \\
   \includegraphics[valign=c,width=0.3\linewidth]{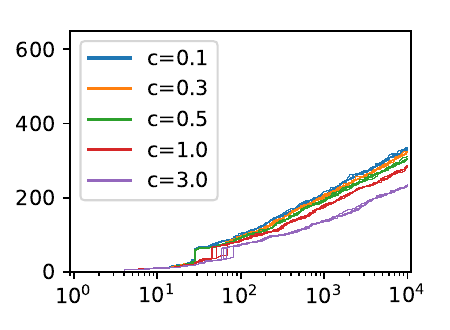}&
   \includegraphics[valign=c,width=0.3\linewidth]{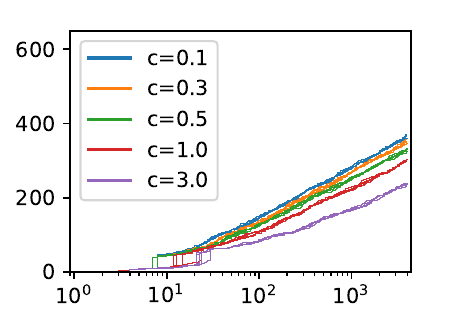}&
   \includegraphics[valign=c,width=0.3\linewidth]{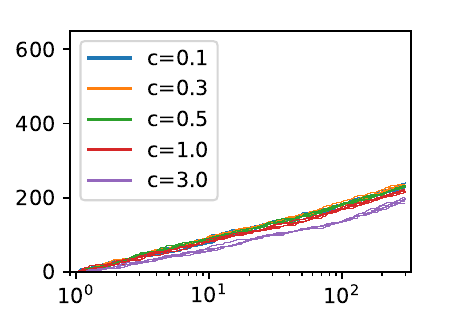}&
   \rotatebox[origin=c]{-90}{$\gc$}
  \end{tabular}
 \caption{
 The cumulative histogram of the number of problem instances solved ($y$-axis)
 below a certain number of node evaluations/expansions/elapsed time ($x$-axis, 10,000 nodes maximum, 4,000 nodes maximum, 300 seconds maximum),
 comparing GUCT with $c\in\braces{0.1,0.3,0.5,1.0,3.0}$.
 Each line represents a random seed.
 }
 \label{fig:different-c}
\end{figure*}

\refigs{fig:different-c}{fig:different-c-01} shows how the hyperparameter $c$
affects the performance of \guct and \guct-01.
In our experiment, \guct tends to perform better with a smaller $c$ value.
As the results with $c=0.1$ tend to be the best,
\reftbl{tbl:main-table-c-0.1} shows a variation of \reftbl{tbl:main-table} with $c=0.1$ instead of $c=0.5$ recommended by \citet{schulte2014balancing}.
This result does not affect the overall trend that \guct and \guct-01 tend to be inferior to \guct-Normal2.

\begin{figure*}[tb]
 \centering
  \begin{tabular}{ccc|c}
   evaluated & expanded & elapsed & $h$ \\\midrule
   \includegraphics[valign=c,width=0.3\linewidth]{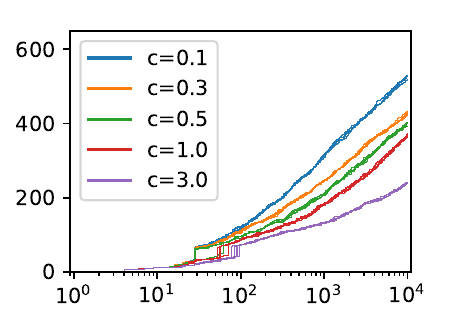}&
   \includegraphics[valign=c,width=0.3\linewidth]{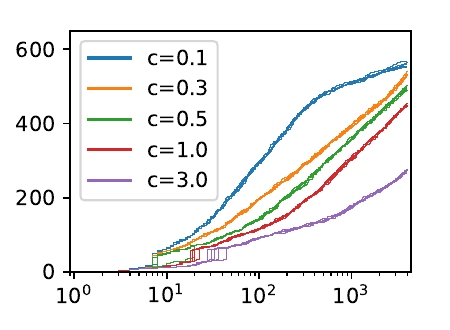}&
   \includegraphics[valign=c,width=0.3\linewidth]{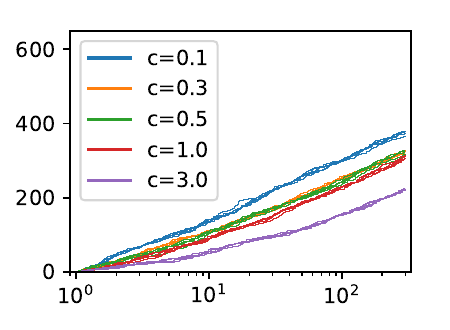}&
   \rotatebox[origin=c]{-90}{$\ff$}
       \\
   \includegraphics[valign=c,width=0.3\linewidth]{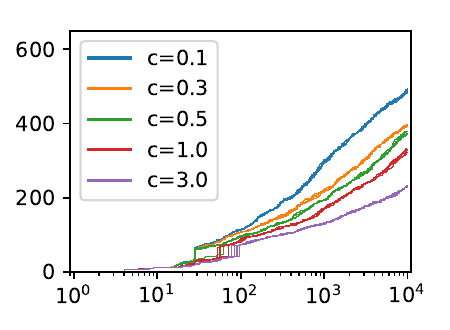}&
   \includegraphics[valign=c,width=0.3\linewidth]{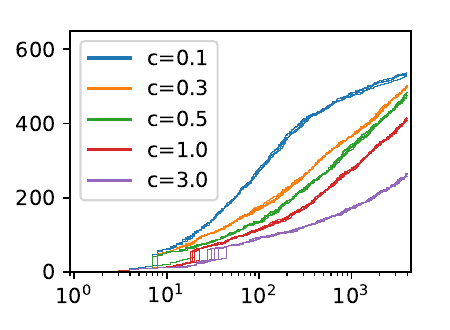}&
   \includegraphics[valign=c,width=0.3\linewidth]{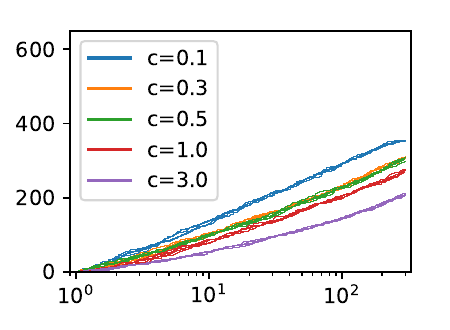}&
   \rotatebox[origin=c]{-90}{$\ad$}
       \\
   \includegraphics[valign=c,width=0.3\linewidth]{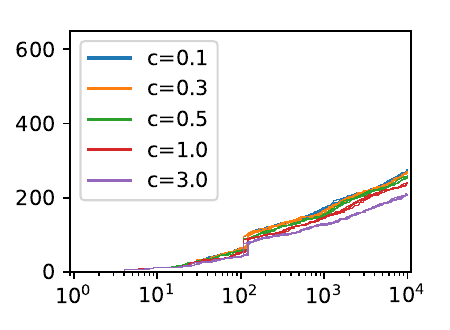}&
   \includegraphics[valign=c,width=0.3\linewidth]{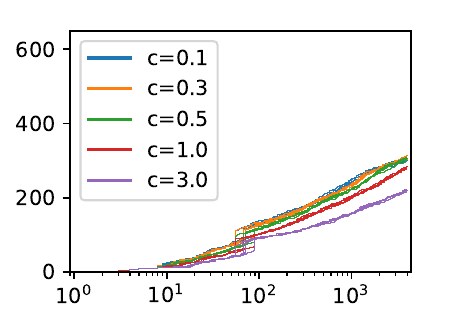}&
   \includegraphics[valign=c,width=0.3\linewidth]{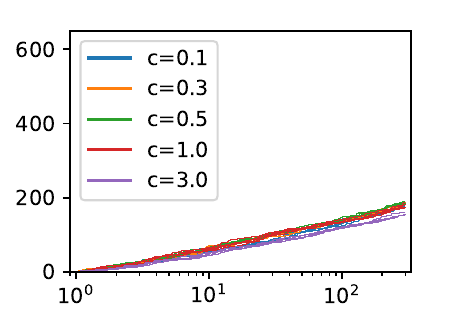}&
   \rotatebox[origin=c]{-90}{$\hmax$}
       \\
   \includegraphics[valign=c,width=0.3\linewidth]{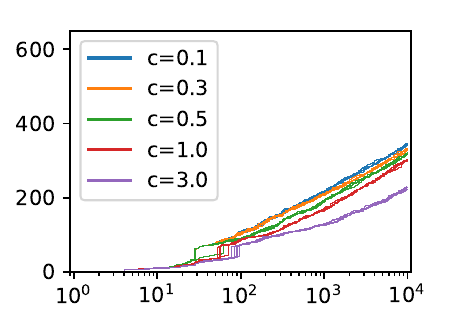}&
   \includegraphics[valign=c,width=0.3\linewidth]{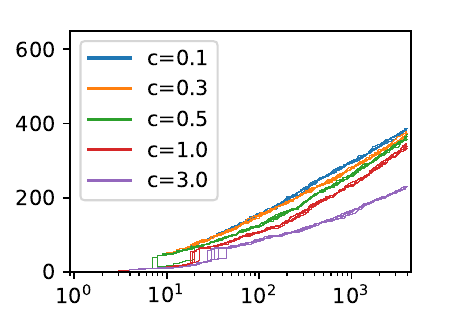}&
   \includegraphics[valign=c,width=0.3\linewidth]{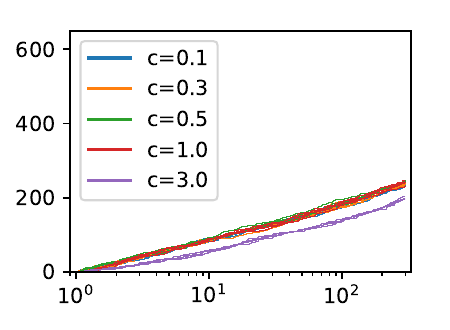}&
   \rotatebox[origin=c]{-90}{$\gc$}
  \end{tabular}
 \caption{
 The cumulative histogram of the number of problem instances solved ($y$-axis)
 below a certain number of node evaluations/expansions/elapsed time ($x$-axis, 10,000 nodes maximum, 4,000 nodes maximum, 300 seconds maximum),
 comparing GUCT-01 with $c\in\braces{0.1,0.3,0.5,1.0,3.0}$.
 Each line represents a random seed.
 }
 \label{fig:different-c-01}
\end{figure*}

\begin{table*}[tb]
 \centering
\begin{adjustbox}{width=\linewidth}
  \begin{tabular}{l*{7}{cc}}
  \toprule
  \multicolumn{1}{r}{$h=$}
  & \multicolumn{2}{c}{$\ff$}
  & \multicolumn{2}{c}{$\ad$}
  & \multicolumn{2}{c}{$\hmax$}
  & \multicolumn{2}{c}{$\gc$}
  & \multicolumn{2}{c}{$\ff$+PO}
  & \multicolumn{2}{c}{$\ff$+DE}
  & \multicolumn{2}{c}{$\ff$+DE+PO}
  \\
  \cmidrule(r){2-3}
  \cmidrule(r){4-5}
  \cmidrule(r){6-7}
  \cmidrule(r){8-9}
  \cmidrule(r){10-11}
  \cmidrule(r){12-13}
  \cmidrule(r){14-15}
  \multicolumn{1}{r}{$c=$}
 & 0.1 & 1 & 0.1 & 1 & 0.1 & 1 & 0.1 & 1 & 0.1 & 1 & 0.1 & 1 & 0.1 & 1 \\
\midrule
GUCT & 528.8 & 412.0 & 504.2 & 397.8 & 260.6 & 228.4 & 334.0 & 285.2 & 454.0 & 389.2 & 439.4 & 0.0 & 0.0 & 0.0 \\ 
* & 555.0 & 458.6 & 537.2 & 480.8 & 263.2 & 242.2 & 328.4 & 310.4 & 495.8 & 423.6 & 471.0 & 0.0 & 0.0 & 0.0 \\ 
-01 & 526.2 & 368.0 & 488.6 & 328.8 & 271.4 & 237.4 & 345.0 & 302.4 & 408.4 & 361.2 & 422.6 & 0.0 & 0.0 & 0.0 \\ 
*-01 & 561.8 & 388.0 & 534.0 & 364.4 & 258.6 & 233.4 & 339.6 & 297.6 & 420.6 & 378.2 & 438.8 & 0.0 & 0.0 & 0.0 \\ 
-V & 443.0 & 317.4 & 441.4 & 310.6 & 248.6 & 208.6 & 326.2 & 255.4 & 389.6 & 344.6 & 421.0 & 0.0 & 0.0 & 0.0 \\ 
\midrule
-Normal & - & 283.4 & - & 265.0 & - & 212.0 & - & 233.4 & - & 372.4 & - & 289.0 & - & 381.6 \\ 
*-Normal & - & 318.8 & - & 300.0 & - & 215.2 & - & 246.2 & - & 378.1 & - & 304.4 & - & 386.7 \\ 
-Normal2 & - & \textbf{581.8} & - & 535.8 & - & \textbf{316.6} & - & \textbf{379.0} & - & \textbf{621.0} & - & \textbf{518.0} & - & \textbf{578.0} \\ 
*-Normal2 & - & 567.2 & - & 533.8 & - & 263.0 & - & 341.0 & - & 618.0 & - & 511.4 & - & 567.8 \\ 
TTTS-Normal & - & 181.0 & - & 180.0 & - & 171.4 & - & 170.8 & - & 151.0 & - & 180.6 & - & 150.6 \\ 
TTTS-Normal* & - & 189.4 & - & 186.4 & - & 177.4 & - & 174.4 & - & 159.4 & - & 185.8 & - & 155.8 \\ 
\midrule
\multicolumn{2}{l}{GBFS(Pyperplan/FastDownward)} & 538/539 & - & 518/517 & - & 224/226 & - & 354/349 & - & $\dagger$/539 & - & 489/$\ddagger$ & - & $\dagger$/$\ddagger$ \\ 
Softmin-Type(h) & - & 576.0 & - & \textbf{542.6} & - & 297.2 & - & 357.6 & - & 575.8 & - & $\ddagger$ & - & $\ddagger$ \\ 
  \bottomrule
 \end{tabular}
\end{adjustbox}
\caption{
 A variation of \reftbl{tbl:main-table}, but $c=0.5$ is replaced by $c=0.1$.
 The best results are unchanged, and configurations with a hyperparameter $c$ never outperformed GUCT-Normal2
 except one case with GUCT* + $\ad$.
 $\dagger$: Data missing due to the lack of support of PO for GBFS in Pyperplan.
 $\ddagger$: Data missing because DE in Fast Downward measures node evaluations differently.
 }
 \label{tbl:main-table-c-0.1}
\end{table*}

\subsection{Solution Quality}

\begin{figure*}[tb]
 \centering
 \includegraphics[width=0.32\linewidth]{img___static___results___scatter2-solution_length___NormalizedUCT-True-hff-False-10-0.2-1.0-0.5-False__algorithm-legend.pdf}
 \includegraphics[width=0.32\linewidth]{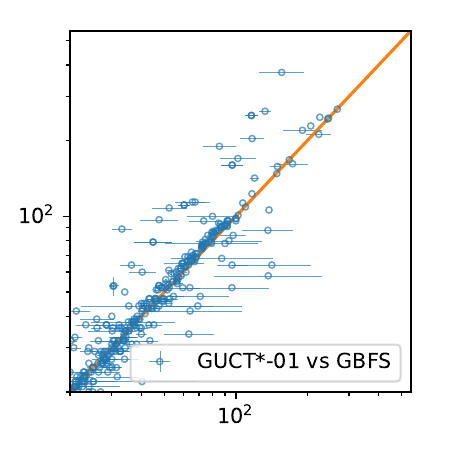}
 \includegraphics[width=0.32\linewidth]{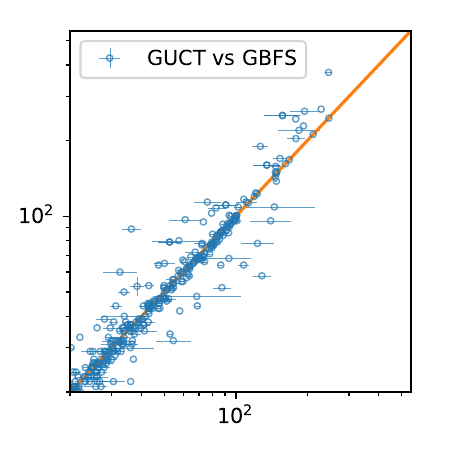}
 \includegraphics[width=0.32\linewidth]{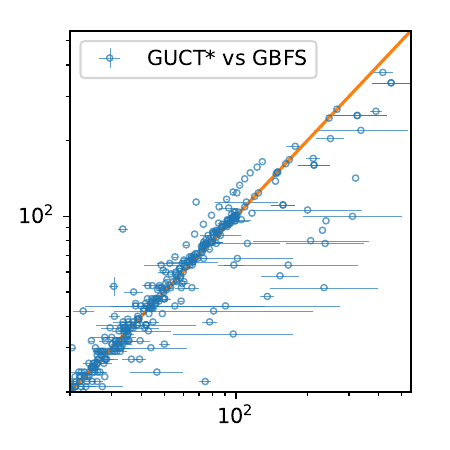}
 \includegraphics[width=0.32\linewidth]{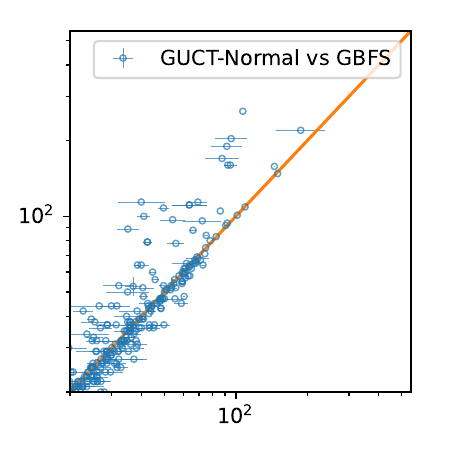}
 \includegraphics[width=0.32\linewidth]{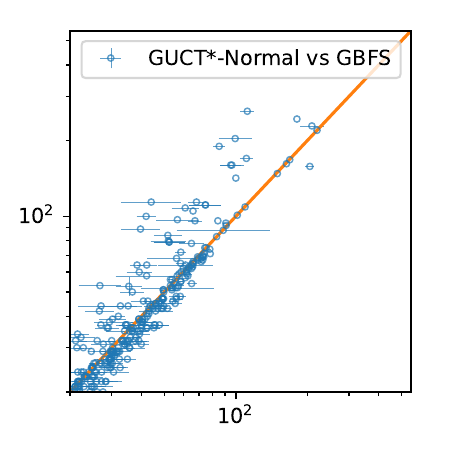}
 \includegraphics[width=0.32\linewidth]{img___static___results___scatter2-solution_length___UCTNormal2-True-hff-False-10-0.2-1.0-0.5-False__algorithm-legend.pdf}
 \includegraphics[width=0.32\linewidth]{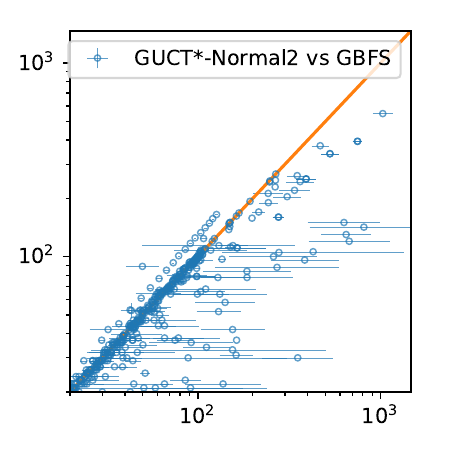}
 \includegraphics[width=0.32\linewidth]{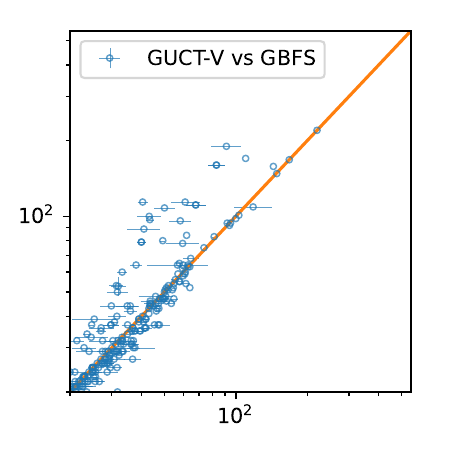}
 \caption{
 Comparing the length of solutions found by \guct-based algorithms ($x$-axis)
 against those by the baseline \gbfs ($y$-axis) using $\ff$.
 }
 \label{fig:solution-length-hff}
\end{figure*}

\refigs{fig:solution-length-hff}{fig:solution-length-gc} shows the complete set of plots for the solution quality.

\begin{figure*}[tb]
 \centering
 \includegraphics[width=0.32\linewidth]{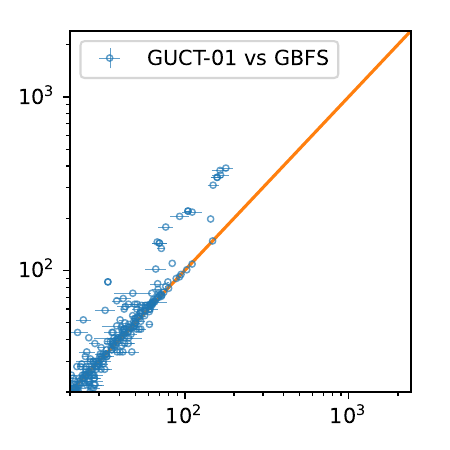}
 \includegraphics[width=0.32\linewidth]{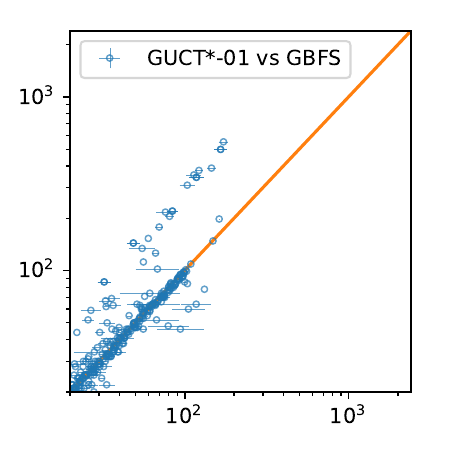}
 \includegraphics[width=0.32\linewidth]{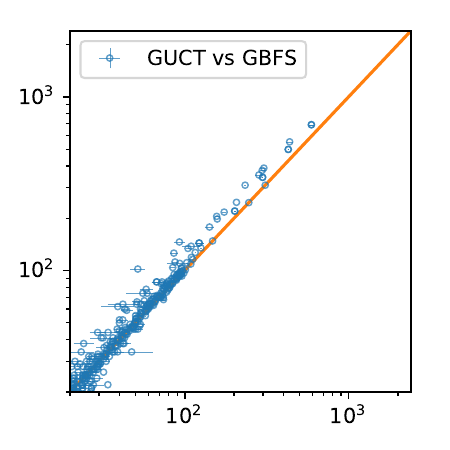}
 \includegraphics[width=0.32\linewidth]{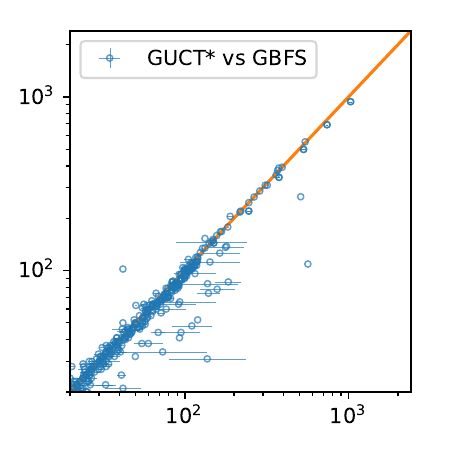}
 \includegraphics[width=0.32\linewidth]{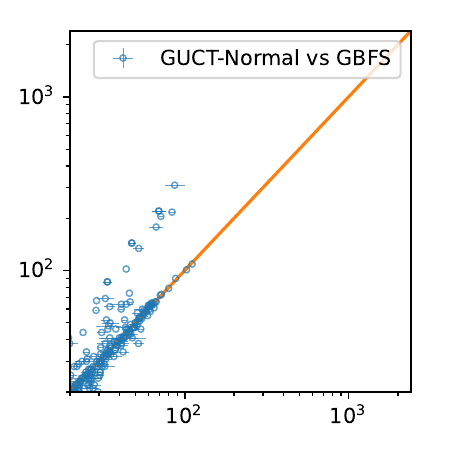}
 \includegraphics[width=0.32\linewidth]{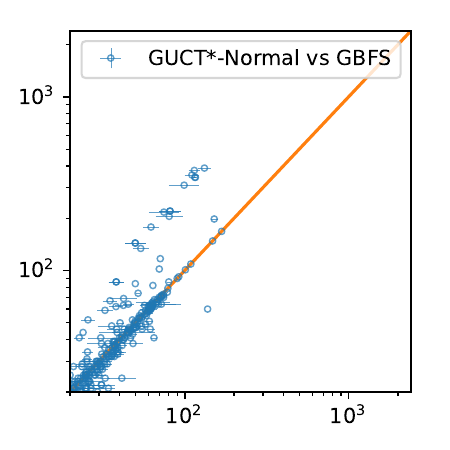}
 \includegraphics[width=0.32\linewidth]{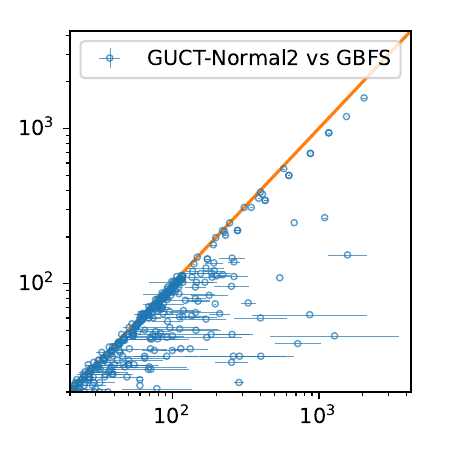}
 \includegraphics[width=0.32\linewidth]{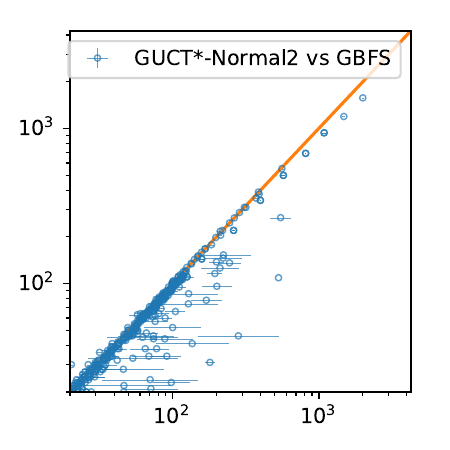}
 \includegraphics[width=0.32\linewidth]{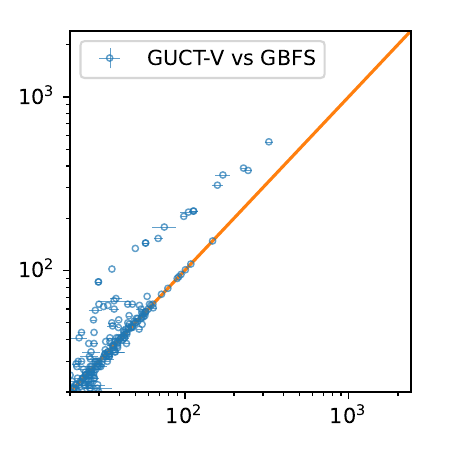}
 \caption{
 Comparing the length of solutions found by \guct-based algorithms ($x$-axis)
 against those by the baseline \gbfs ($y$-axis) using $\ad$.
 }
 \label{fig:solution-length-hadd}
\end{figure*}

\begin{figure*}[tb]
 \centering
 \includegraphics[width=0.32\linewidth]{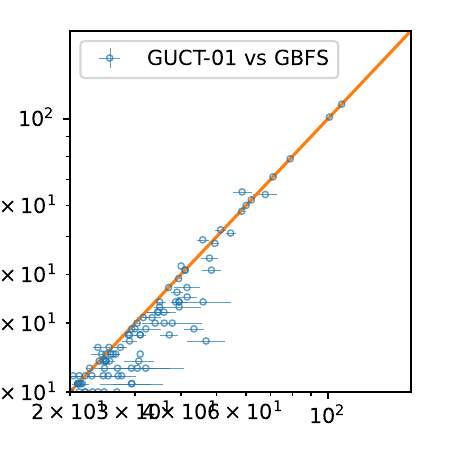}
 \includegraphics[width=0.32\linewidth]{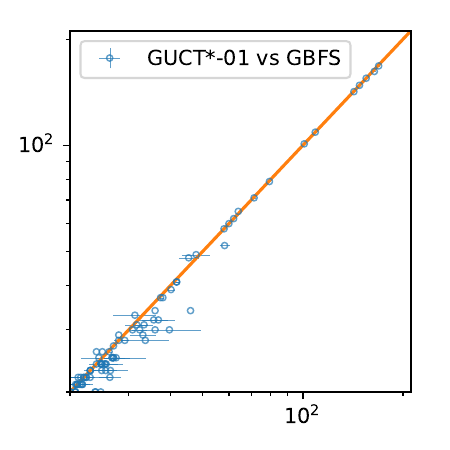}
 \includegraphics[width=0.32\linewidth]{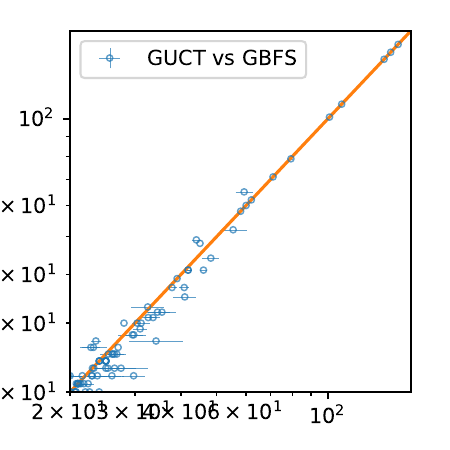}
 \includegraphics[width=0.32\linewidth]{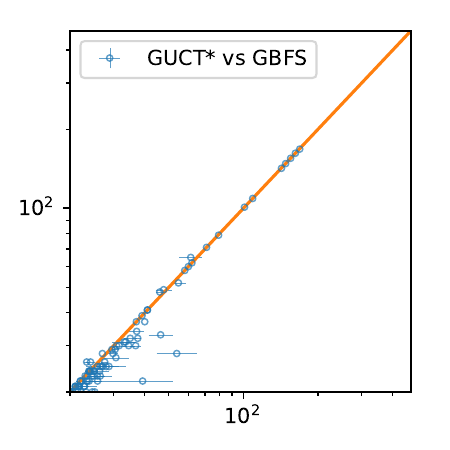}
 \includegraphics[width=0.32\linewidth]{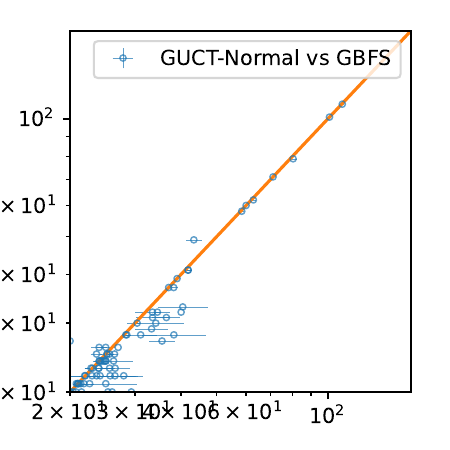}
 \includegraphics[width=0.32\linewidth]{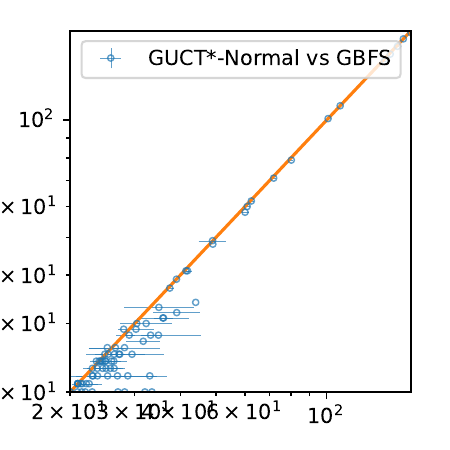}
 \includegraphics[width=0.32\linewidth]{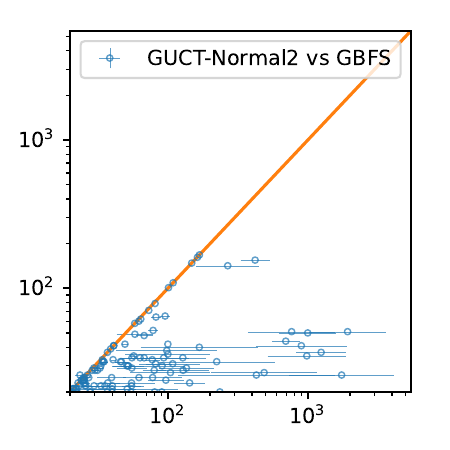}
 \includegraphics[width=0.32\linewidth]{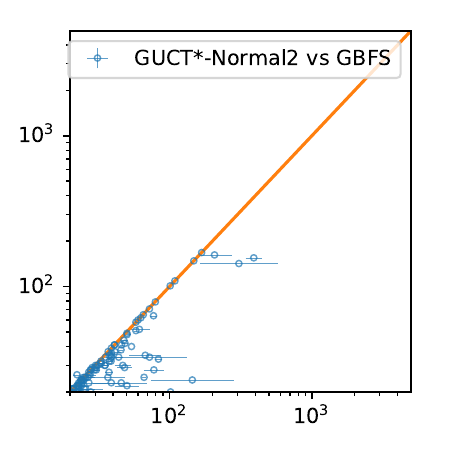}
 \includegraphics[width=0.32\linewidth]{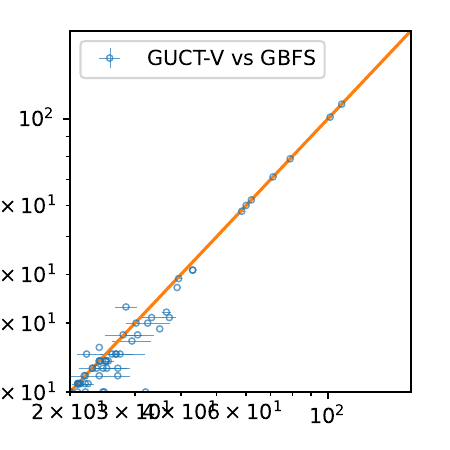}
 \caption{
 Comparing the length of solutions found by \guct-based algorithms ($x$-axis)
 against those by the baseline \gbfs ($y$-axis) using $\hmax$.
 }
 \label{fig:solution-length-hmax}
\end{figure*}

\begin{figure*}[tb]
 \centering
 \includegraphics[width=0.32\linewidth]{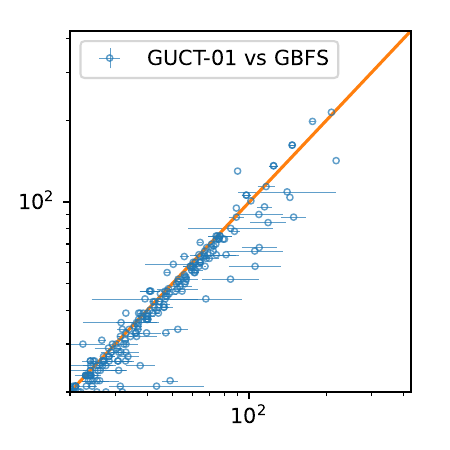}
 \includegraphics[width=0.32\linewidth]{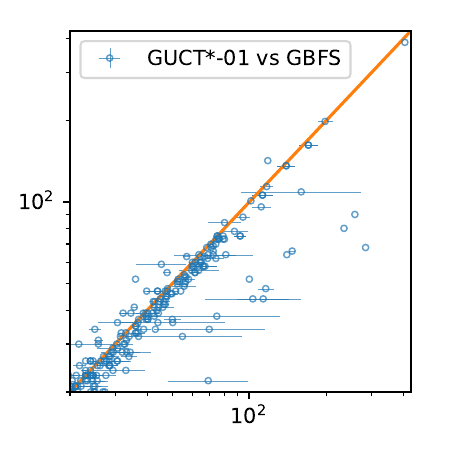}
 \includegraphics[width=0.32\linewidth]{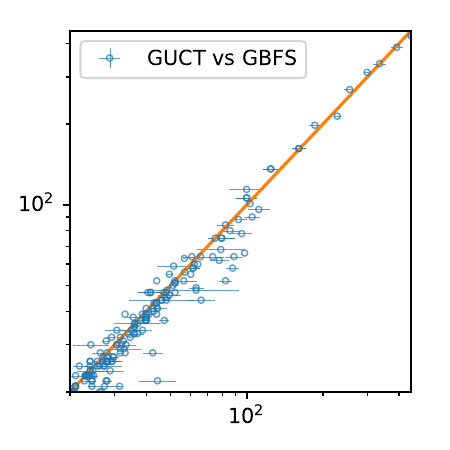}
 \includegraphics[width=0.32\linewidth]{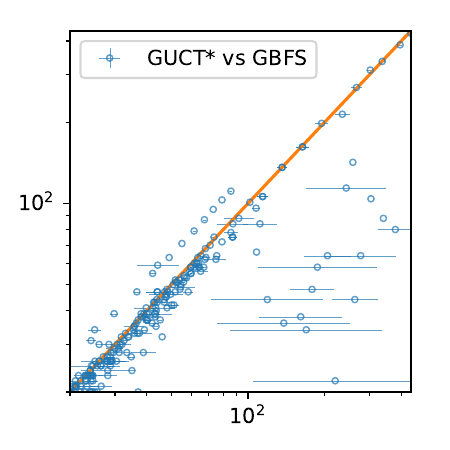}
 \includegraphics[width=0.32\linewidth]{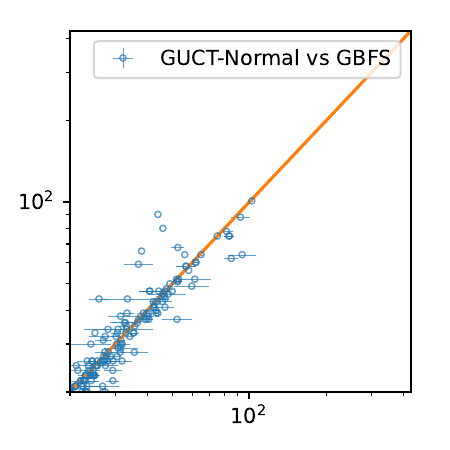}
 \includegraphics[width=0.32\linewidth]{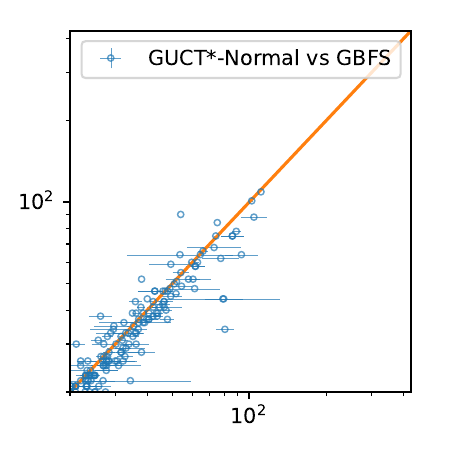}
 \includegraphics[width=0.32\linewidth]{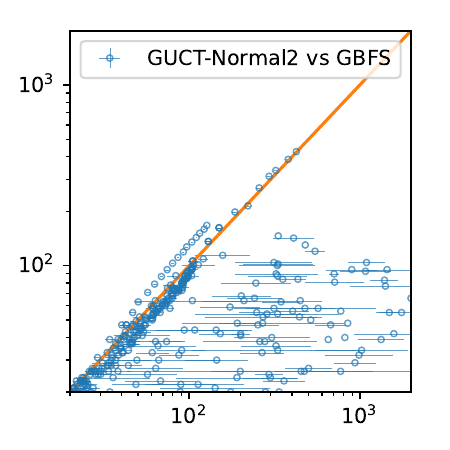}
 \includegraphics[width=0.32\linewidth]{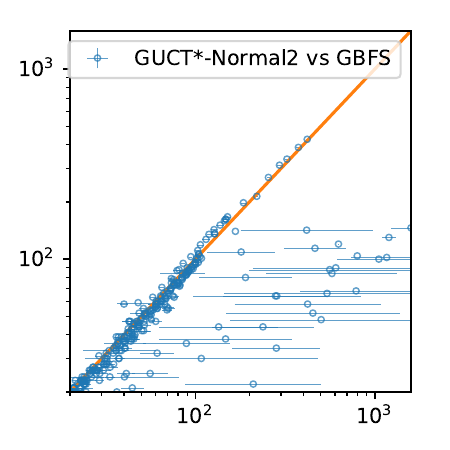}
 \includegraphics[width=0.32\linewidth]{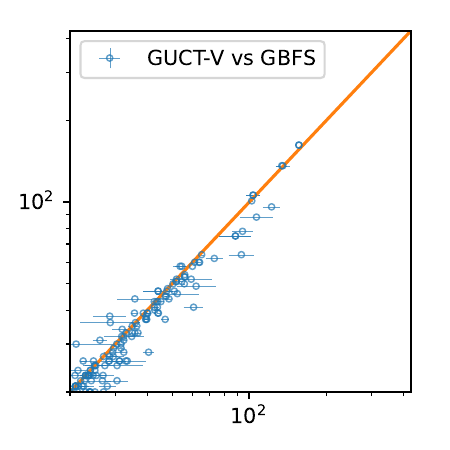}
 \caption{
 Comparing the length of solutions found by \guct-based algorithms ($x$-axis)
 against those by the baseline \gbfs ($y$-axis) using $\gc$.
 }
 \label{fig:solution-length-gc}
\end{figure*}

\subsection{Domain-wise comparison}
\label{sec:domain-wise}

\begin{figure*}[tb]
 \centering
 \includegraphics[width=0.5\linewidth]{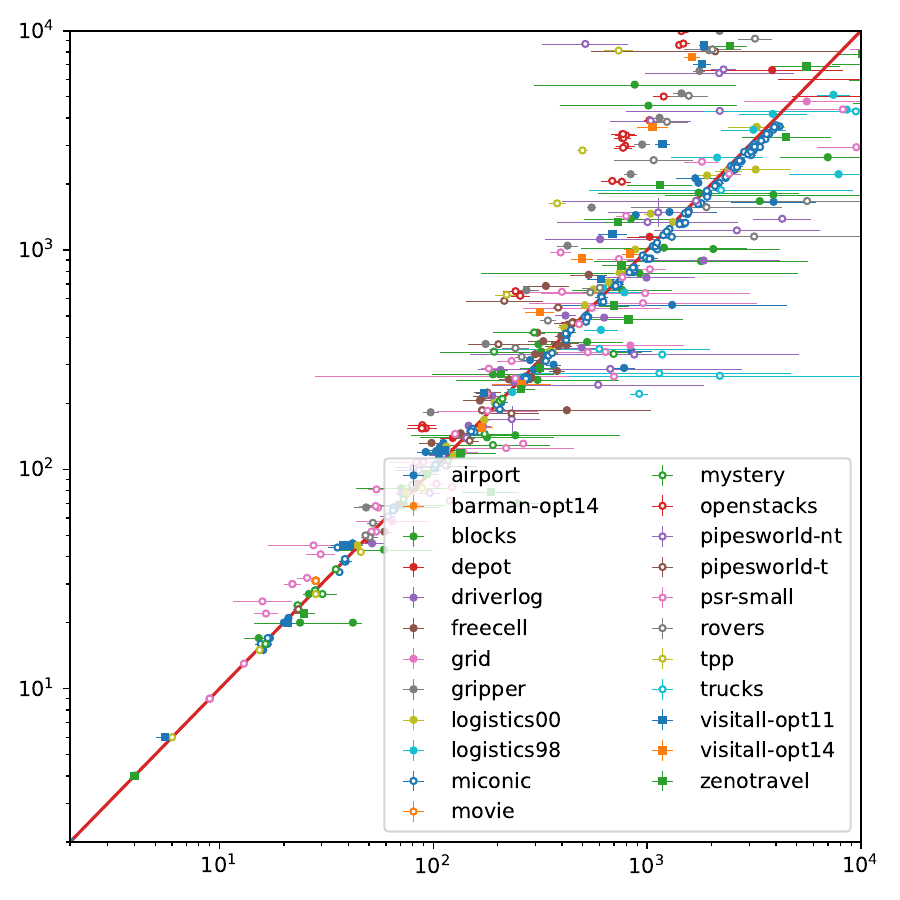}
 \caption{
 The scatter plot of node evaluations for GBFS ($y$-axis) and GUCT-Normal2 ($x$-axis) for instances solved by both.
 The error bar shows the fluctuation due to 5 random seeds (applicable only to GUCT-Normal2).
 }
 \label{fig:evaluation-scatter-uctnormal2-gbfs}
\end{figure*}

In order to see the effect of domains on the performance,
\reftbl{tbl:domainwise-normal2-gbfs-softmin} shows the domain-wise coverage comparison
between GUCT-Normal2, GBFS, and Softmin-Type(h)
evaluated with $\ff$ heuristic, $\ad$ heuristic, and $\ff$+PO configuration.
In addition, 
in \refigs{fig:evaluation-scatter-uctnormal2-gbfs}{fig:evaluation-scatter-uctnormal2-softmin}
we plotted the number of node evaluations for GUCT-Normal2, GBFS, and Softmin-Type(h)
in individual instances solved by both configurations
and colored the points according to the domain.

\begin{figure*}[tb]
 \centering
 \includegraphics[width=0.5\linewidth]{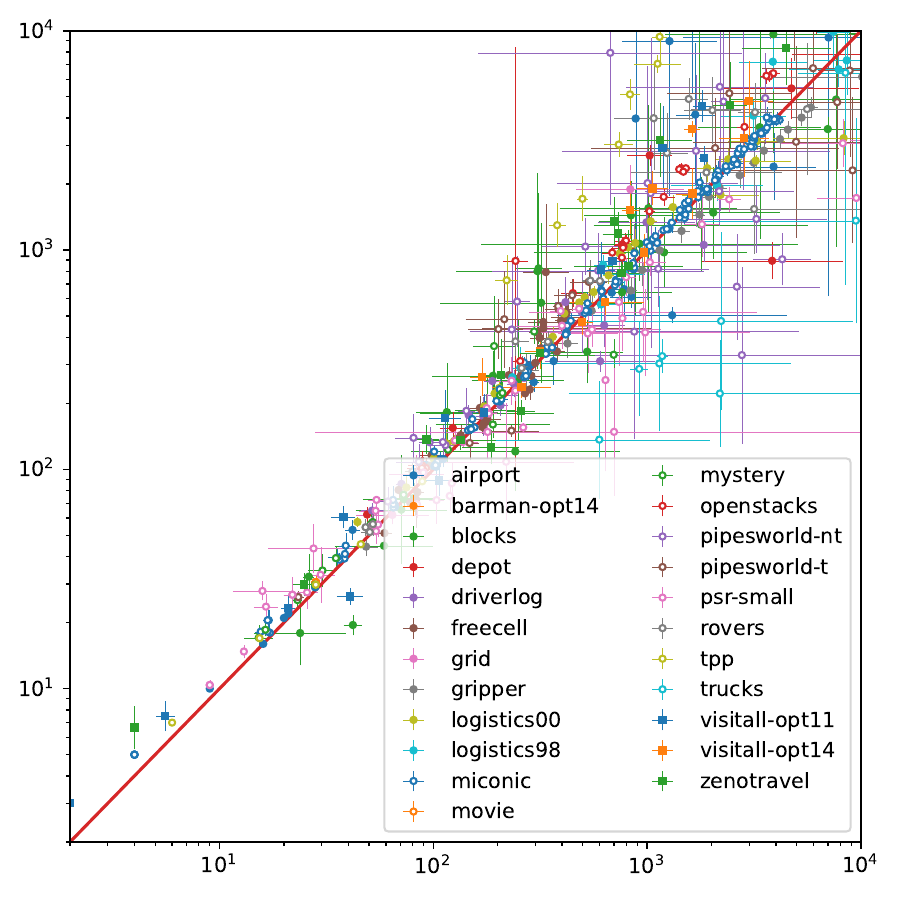}
 \caption{
 The scatter plot of node evaluations for Softmin-Type(h) ($y$-axis) and GUCT-Normal2 ($x$-axis) for instances solved by both.
 The error bar shows the fluctuation due to 5 random seeds.
 }
 \label{fig:evaluation-scatter-uctnormal2-softmin}
\end{figure*}

\begin{table*}[htbp]
\begin{adjustbox}{width=\linewidth}
\begin{tabular}{lrrrrrrrrr}
\toprule
 & \multicolumn{3}{c}{$\ff$} & \multicolumn{3}{c}{$\ad$} & \multicolumn{3}{c}{$\ff$+PO} \\ 
\cmidrule(r){2-4}
\cmidrule(r){5-7}
\cmidrule(r){8-10}
domain & -Normal2 & GBFS(Pp) & Softmin & -Normal2 & GBFS(Pp) & Softmin & -Normal2 & GBFS(FD) & Softmin \\ 
\midrule
airport        & 22            & \textbf{27} & 26.8          & \textbf{25}   & \textbf{25} & 24.8          & 21.4          & \textbf{27} & 26.8          \\ 
barman-opt14   & \textbf{6}    & 0           & 0             & \textbf{2.2}  & 0           & 0             & \textbf{14}   & 0           & 0             \\ 
blocks         & 29.8          & \textbf{31} & 30.4          & 34.4          & \textbf{35} & 34.8          & \textbf{34.8} & 28          & 30.4          \\ 
depot          & \textbf{8.8}  & 7           & 8.2           & 6.2           & 6           & \textbf{6.8}  & \textbf{12}   & 6           & 8.2           \\ 
driverlog      & \textbf{14.2} & 14          & 14            & 14.4          & 14          & \textbf{14.6} & \textbf{16.6} & 14          & 14            \\ 
freecell       & 34            & 34          & 34            & 34            & 34          & 34            & 34            & 34          & 34            \\ 
grid           & 3.6           & \textbf{4}  & 3             & 2             & 2           & \textbf{3}    & \textbf{5}    & 3           & 3             \\ 
gripper        & \textbf{20}   & 13          & \textbf{20}   & 10            & \textbf{20} & \textbf{20}   & \textbf{20}   & 13          & \textbf{20}   \\ 
logistics00    & 27.4          & 26          & \textbf{28}   & 28            & 28          & 28            & 28            & 28          & 28            \\ 
logistics98    & 10.6          & \textbf{12} & 11            & 9.6           & \textbf{11} & 9.8           & \textbf{15}   & 11          & 11            \\ 
miconic        & 150           & 150         & 150           & 150           & 150         & 150           & 150           & 150         & 150           \\ 
movie          & 30            & 30          & 30            & 30            & 30          & 30            & 0             & \textbf{30} & \textbf{30}   \\ 
mystery        & 15.2          & 16          & \textbf{16.4} & 15.4          & \textbf{17} & \textbf{17}   & 12.8          & 16          & \textbf{16.4} \\ 
openstacks     & \textbf{27}   & 21          & 25.8          & 5             & \textbf{7}  & 5             & \textbf{27}   & 22          & 25.8          \\ 
pipesworld-nt  & 26.4          & 23          & \textbf{28}   & 20.4          & 12          & \textbf{27.8} & \textbf{41.2} & 21          & 28            \\ 
pipesworld-t   & 12.8          & 9           & \textbf{15.6} & \textbf{12.8} & 9           & 11.4          & \textbf{19.8} & 11          & 15.6          \\ 
psr-small      & 44.8          & 47          & \textbf{48.6} & 40.4          & 43          & \textbf{43.4} & \textbf{50}   & 47          & 48.6          \\ 
rovers         & \textbf{18.8} & 17          & 18            & \textbf{17}   & 13          & 16.8          & \textbf{29.8} & 17          & 17.8          \\ 
tpp            & \textbf{23.6} & 9           & 11.4          & \textbf{24.4} & 9           & 12.6          & \textbf{29.6} & 9           & 11.4          \\ 
trucks         & 6.4           & 8           & \textbf{9.6}  & 5.4           & \textbf{7}  & 6.6           & 6             & 8           & \textbf{9.6}  \\ 
visitall-opt11 & \textbf{19.6} & 14          & 19.2          & \textbf{20}   & 19          & 19            & \textbf{20}   & 19          & 19.2          \\ 
visitall-opt14 & \textbf{13.4} & 7           & 12.4          & \textbf{13}   & 12          & 12            & \textbf{14}   & 12          & 12.4          \\ 
zenotravel     & 17.4          & \textbf{19} & 15.6          & \textbf{16.2} & 15          & 15.2          & \textbf{20}   & 13          & 15.6          \\ 
\midrule 
Total & \textbf{581.8} & 538 & 576 & 535.8 & 518 & \textbf{542.6} & \textbf{621} & 539  & 575.8 \\ 
\bottomrule
\end{tabular}
\end{adjustbox}
\caption{
 Domain-wise coverage for $\ff$ heuristic, $\ad$ heuristic, and $\ff$+PO configuration.
 (Pp) stands for Pyperplan, and (FD) stands for Fast Downward.
 Best results in bold except all ties.
 GUCT-Normal2's failure in movie could be a bug, but we do not consider it an issue since it only disadvantages our proposed method.
}
\label{tbl:domainwise-normal2-gbfs-softmin}
\end{table*}

\fontsize{9.5pt}{10.5pt}
\selectfont

\end{document}